\documentclass[10pt,twocolumn,letterpaper]{article}
\usepackage{authblk}

\usepackage[pagenumbers]{cvpr} 

\usepackage{graphicx}
\usepackage{amsmath}
\usepackage{amssymb}
\usepackage{booktabs}
\usepackage{lipsum}
\usepackage{ulem}
\usepackage{makecell}

\newcommand{\reducespace}{\vspace{-3pt}}

\newif\ifdrafting
\draftingtrue 

\ifdrafting
    \newcommand{\kn}[1]{\textcolor{cyan}{[KN: #1]}}
    \newcommand{\sg}[1]{\textcolor{green}{[Shalini: #1]}}
    \newcommand{\gw}[1]{\textcolor{red}{[GW: #1]}}
    \newcommand{\mm}[1]{\textcolor{magenta}{[MM: #1]}}
    
    \newcommand{\ec}[1]{\textcolor{blue}{[EC: #1]}}
    \newcommand{\mc}[1]{\textcolor{blue}{MC: #1}}

    \newcommand{\og}[1]{\textcolor{blue}{[OG: #1]}}
    \newcommand{\tk}[1]{\textcolor{red}{[TK: #1]}}
\else
    \newcommand{\kn}[1]{}
    \newcommand{\sg}[1]{}
    \newcommand{\gw}[1]{}
    \newcommand{\mm}[1]{}
    \newcommand{\ec}[1]{}
    \newcommand{\mc}[1]{}
    \newcommand{\og}[1]{}
    \newcommand{\tk}[1]{}
    \newcommand{\edit}[1]{}
\fi

\newcommand{\myparagraph}[1]{\vspace{-8pt}\paragraph{#1}}

\usepackage[pagebackref,breaklinks,colorlinks]{hyperref}
\usepackage{enumitem}

\usepackage[capitalize]{cleveref}
\crefname{section}{Sec.}{Secs.}
\Crefname{section}{Section}{Sections}
\Crefname{table}{Table}{Tables}
\crefname{table}{Tab.}{Tabs.}

\def\cvprPaperID{3308} 
\def\confName{CVPR}
\def\confYear{2022}

\begin{document}

\title{Efficient Geometry-aware 3D Generative Adversarial Networks}

\makeatletter
    \renewcommand\AB@affilsepx{ \hphantom{---} \protect\Affilfont}
\makeatother

\newcommand*\samethanks[1][\value{footnote}]{\footnotemark[#1]}

\author[1,2]{Eric R. Chan \thanks{Equal contribution.}\thanks{Part of the work was done during an internship at NVIDIA.}}
\author[1]{Connor Z. Lin\samethanks[1]}
\author[1]{Matthew A. Chan\samethanks[1]}
\author[2]{Koki Nagano\samethanks[1]}
\author[1]{Boxiao Pan}
\author[2]{Shalini De Mello}
\author[2]{Orazio Gallo}
\author[1]{Leonidas Guibas}
\author[2]{Jonathan Tremblay}
\author[2]{Sameh Khamis}
\author[2]{Tero Karras}
\author[1]{Gordon Wetzstein}
\affil[1]{Stanford University}
\affil[2]{NVIDIA}

\maketitle
 
\global\csname @topnum\endcsname 0
\global\csname @botnum\endcsname 0

\begin{abstract}
Unsupervised generation of high-quality multi-view-consistent images and 3D shapes using only collections of single-view 2D photographs has been a long-standing challenge. Existing 3D GANs are either compute-intensive or make approximations that are not 3D-consistent; the former limits quality and resolution of the generated images and the latter adversely affects multi-view consistency and shape quality. In this work, we improve the computational efficiency and image quality of 3D GANs without overly relying on these approximations. We introduce an expressive hybrid explicit-implicit network architecture that, together with other design choices, synthesizes not only high-resolution multi-view-consistent images in real time but also produces high-quality 3D geometry. By decoupling feature generation and neural rendering, our framework is able to leverage state-of-the-art 2D CNN generators, such as StyleGAN2, and inherit their efficiency and expressiveness. We demonstrate state-of-the-art 3D-aware synthesis with FFHQ and AFHQ Cats, among other experiments.
\end{abstract}

\newcommand\blfootnote[1]{%
  \begingroup
  \renewcommand\thefootnote{}\footnote{#1}%
  \addtocounter{footnote}{-1}%
  \endgroup
}

\blfootnote{Project page: \url{https://github.com/NVlabs/eg3d}}



\begin{figure}
    \centering
    \includegraphics[width=\linewidth]{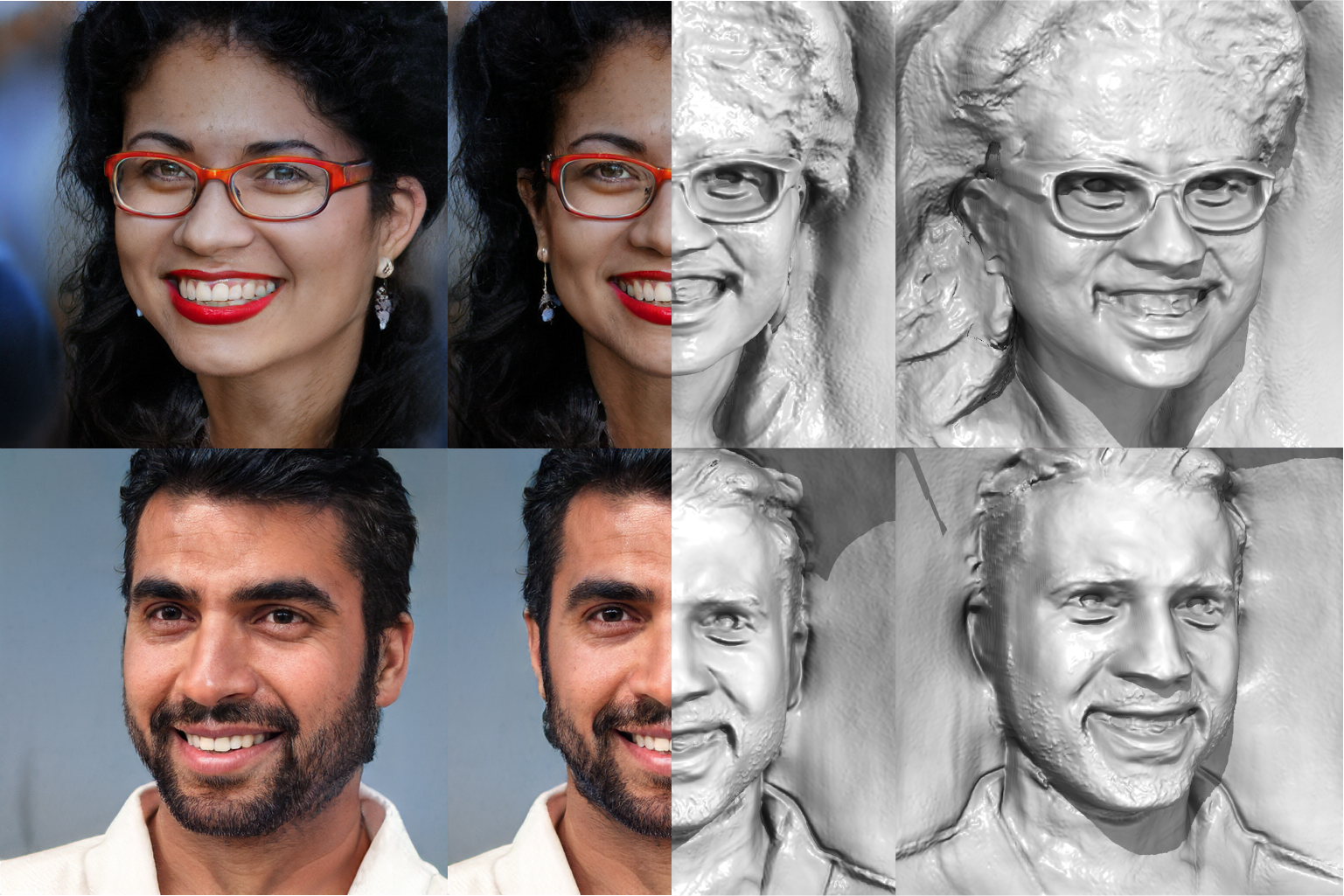}
    \caption{
    Our 3D GAN enables synthesis of scenes, producing high-quality, multi-view-consistent renderings and detailed geometry. Our approach trains from a collection of 2D images without target-specific shape priors, ground truth 3D scans, or multi-view supervision. Please see the accompanying video for more results.}
    \label{fig:teaser}
\end{figure}

\section{Introduction}
\reducespace
\label{sec:intro}

Generative adversarial networks (GANs) have seen immense progress, with recent models capable of generating high-resolution, photorealistic images indistinguishable from real photographs~\cite{karras2019style,Karras2020stylegan2,Karras2021}. 
Current state-of-the-art GANs, however, operate in 2D only and do not explicitly model the underlying 3D scenes.


Recent work on 3D-aware GANs has begun to tackle the problem of multi-view-consistent image synthesis and, to a lesser extent, extraction of 3D shapes without being supervised on geometry or multi-view image collections. 
%
However, the image quality and resolution of existing 3D GANs have lagged far behind those of 2D GANs. Furthermore, their 3D reconstruction quality, so far, leaves much to be desired. 
One of the primary reasons for this gap is the computational inefficiency of previously employed 3D generators and neural rendering architectures.

In contrast to 2D GANs, 3D GANs rely on a combination of a 3D-structure-aware inductive bias in the generator network architecture and a neural rendering engine that aims at providing view-consistent results. The inductive bias can be modeled using explicit voxel grids~\cite{wu2016learning,Gadelha:2017,zhu2018visual,henzler2019platonicgan,hologan,nguyen2020blockgan} or neural implicit representations~\cite{hologan,graf,chan2020pi,Niemeyer2020GIRAFFE}.
%
While successful in single-scene ``overfitting'' scenarios, neither of these representations is suitable for training a high-resolution 3D GAN because they are simply too memory inefficient or slow.
Training a 3D GAN requires rendering tens of millions of images, but state-of-the-art neural volume rendering\cite{mildenhall2020nerf} at high-resolutions with these representations is computationally infeasible.
%
%
CNN-based image upsampling networks have been proposed to remedy this~\cite{Niemeyer2020GIRAFFE}, but such an approach sacrifices view consistency and impairs the quality of the learned 3D geometry.

We introduce a novel generator architecture for unsupervised 3D representation learning from a collection of single-view 2D photographs that seeks to improve the computational efficiency of rendering while remaining true to 3D-grounded neural rendering. We achieve this goal with a two-pronged approach. First, we improve the computational efficiency of 3D-grounded rendering with a hybrid explicit--implicit 3D representation that offers significant speed and memory benefits over fully implicit or explicit approaches without compromising on expressiveness. These advantages enable our method to skirt the computational constraints that have limited the rendering resolutions and quality of previous approaches~\cite{graf,chan2020pi} and forced over-reliance on image-space convolutional upsampling~\cite{Niemeyer2020GIRAFFE}. Second, although we use some image-space approximations that stray from the 3D-grounded rendering, we introduce a dual-discrimination strategy that maintains consistency between the neural rendering and our final output to regularize their undesirable view-inconsistent tendencies. Moreover, we introduce pose-based conditioning to our generator, which decouples pose-correlated attributes (e.g., facial expressions) for a multi-view consistent output during inference while faithfully modeling the joint distributions of pose-correlated attributes inherent in the training data. 

As an additional benefit, our framework decouples feature generation from neural rendering, enabling it to directly leverage state-of-the-art 2D CNN-based feature generators, such as StyleGAN2, to generalize over spaces of 3D scenes while also benefiting from 3D multi-view-consistent neural volume rendering. Our approach not only achieves state-of-the-art qualitative and quantitative results for view-consistent 3D-aware image synthesis, but also generates high-quality 3D shapes of the synthesized scenes due to its strong 3D-structure-aware inductive bias (see Fig.~\ref{fig:teaser}).

Our contributions are the following:
\begin{itemize}[noitemsep]
    \item We introduce a tri-plane-based 3D GAN framework, which is both efficient and expressive, to enable high-resolution geometry-aware image synthesis. 
    \item We develop a 3D GAN training strategy that promotes multi-view consistency via dual discrimination and generator pose conditioning while faithfully modeling pose-correlated attribute distributions (e.g., expressions) present in real-world datasets.
    \item We demonstrate state-of-the-art results for unconditional 3D-aware image synthesis on the FFHQ and AFHQ Cats datasets along with high-quality 3D geometry learned entirely from 2D in-the-wild images. 
\end{itemize}

\section{Related work}
\reducespace
\label{sec:related_works}
\paragraph{Neural scene representation and rendering.}

\begin{figure}[t!]
	\centering
		\includegraphics[width=\columnwidth]{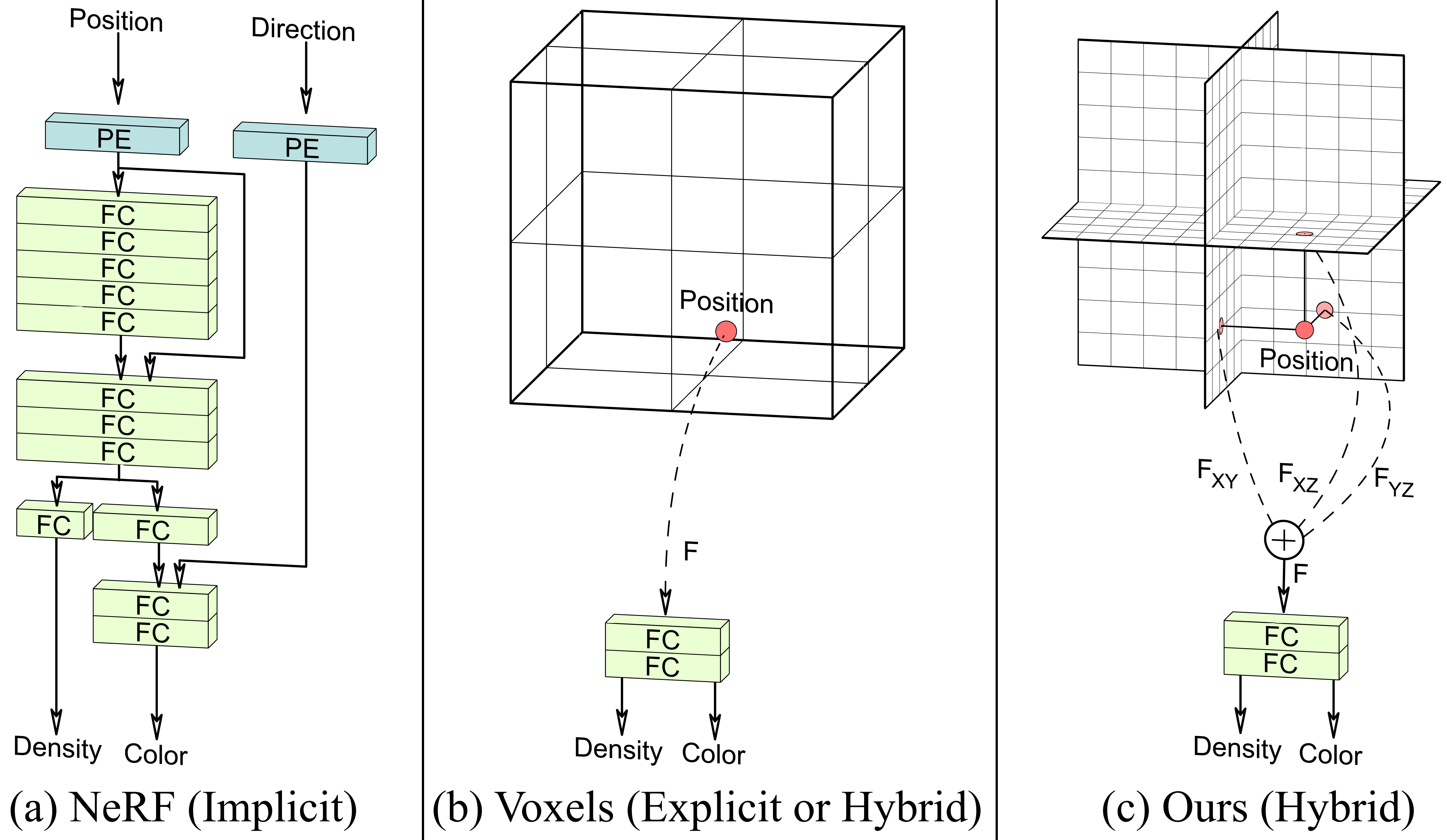}
		\caption{Neural implicit representations use fully connected layers (FC) with positional encoding (PE) to represent a scene, which can be slow to query (a). Explicit voxel grids or hybrid variants using small implicit decoders are fast to query, but scale poorly with resolution (b). Our hybrid explicit--implicit tri-plane representation (c) is fast and scales efficiently with resolution, enabling greater detail for equal capacity.}
		\label{fig:hybrid_representation}
\end{figure}

Emerging neural scene representations use differentiable 3D-aware representations~\cite{eslami2018neural,park2019deepsdf,mescheder2019occupancy,michalkiewicz2019implicit,chen2019learning,atzmon2019sal,gropp2020implicit,davies2020overfit,chabra2020deep,takikawa2021nglod} that can be optimized using 2D multi-view images via neural rendering~\cite{sitzmann2019srns,liu2019learning,mildenhall2020nerf,martinbrualla2020nerfw,Niemeyer2020CVPR,pumarola2020d,srinivasan2020nerv,zhang2020nerf,neff2021donerf,liu2020dist,yariv2020multiview,lindell2020autoint,liu2020neural,jiang2020sdfdiff,kellnhofer2021neural,oechsle2021unisurf,yu2021plenoctrees,hedman2021snerg,garbin2021fastnerf}. 
Explicit representations, such as discrete voxel grids (Fig.~\ref{fig:hybrid_representation}b), are fast to evaluate but often incur heavy memory overheads, making them difficult to scale to high resolutions or complex scenes~\cite{sitzmann2019deepvoxels,Lombardi:2019}. 
Implicit representations, or coordinate networks (Fig.~\ref{fig:hybrid_representation}a), offer potential advantages in memory efficiency and scene complexity compared to discrete voxel grids by representing a scene as a continuous function (e.g.,~\cite{park2019deepsdf,mescheder2019occupancy,tancik2020fourier,sitzmann2020siren, mildenhall2020nerf}). 
In practice, these implicit architectures use large fully connected networks that are slow to evaluate as each query requires a full pass through the network.
Therefore, fully explicit and implicit representations provide complementary benefits.

Local implicit representations~\cite{chabra2020deep,jiang2020local,chen2021learning,Reiser2021ICCV} and hybrid explicit--implicit representations~\cite{devries2021unconstrained, liu2020neural,peng2020convolutional,martel2021acorn} combine the benefits of both types of representations by offering computationally and memory-efficient architectures. 
Inspired by these ideas, we design a new hybrid explicit--implicit 3D-aware network that uses a memory-efficient tri-plane representation to explicitly store features on axis-aligned planes that are aggregated by a lightweight implicit feature decoder for efficient volume rendering (Fig.~\ref{fig:hybrid_representation}c). 
Our representation bears some resemblance to previous plane-based hybrid architectures~\cite{peng2020convolutional,devries2021unconstrained}, but it is unique in its specific design.  
Our representation is key to enabling the high 3D GAN image quality that we demonstrate through efficient training comparable (in time scales) to modern 2D GANs~\cite{Karras2021}. 

\myparagraph{Generative 3D-aware image synthesis.} 

Generative adversarial networks~\cite{goodfellow2014generative} have recently achieved photorealistic image quality for 2D image synthesis~\cite{radford2016unsupervised,karras2018progressive,karras2019style,Karras2020stylegan2}.
Extending these capabilities to 3D settings has started to gain momentum as well.
Mesh-based approaches build on the most popular primitives used in computer graphics, but lack the expressiveness needed for high-fidelity image generation~\cite{Szabo:2019,Liao2020CVPR}. Voxel-based GANs directly extend the CNN generators used in 2D settings to 3D~\cite{wu2016learning,Gadelha:2017,zhu2018visual,henzler2019platonicgan,hologan,nguyen2020blockgan}. The high memory requirements of voxel grids and the computational burden of 3D convolutions, however, make high-resolution 3D GAN training difficult. Low-resolution 3D volume generation can be remedied with 2D CNN-based image upsampling layers~\cite{Niemeyer2020GIRAFFE}, but without an inductive 3D bias the results often lack view consistency. 
Block-based sparse volume representations overcome some of these issues, but are applicable to mostly empty scenes~\cite{hao2021GANcraft, liu2020neural} and difficult to generalize across scenes. As an alternative, fully implicit representation networks have been proposed for 3D scene generation~\cite{graf,chan2020pi}, but these architectures are slow to query, which makes the GAN training inefficient, limiting the quality and resolution of generated images.


One of the primary insights of our work is that an efficient 3D GAN architecture with 3D-grounded inductive biases is crucial for successfully generating high-resolution view-consistent images and high-quality 3D shapes. Our framework achieves this in several ways. First, unlike most existing 3D GANs, we directly leverage a 2D CNN-based feature generator, i.e., StyleGAN2~\cite{Karras2020stylegan2}, removing the need for inefficient 3D convolutions on explicit voxel grids. Second, our tri-plane representation allows us to leverage neural volume rendering as an inductive bias, but in a much more computationally efficient way than fully implicit 3D networks~\cite{mildenhall2020nerf,graf,chan2020pi}. Similar to~\cite{Niemeyer2020GIRAFFE}, we also employ 2D CNN-based upsampling after neural rendering, but our method introduces dual discrimination to avoid view inconsistencies introduced by the upsampling layers. Unlike existing StyleGAN2-based 2.5D GANs, which generate images and depth maps~\cite{shi2021lifting}, our method works naturally for steep camera angles and in 360$^\circ$ viewing conditions.

The concurrently developed 3D-aware GANs StyleNeRF~\cite{gu2021stylenerf} and CIPS-3D~\cite{Zhou2021CIPS3D} demonstrate impressive image quality. 
The central distinction between these and ours is that while StyleNeRF and CIPS-3D operate primarily in image-space, with less emphasis on the 3D representation, our method operates primarily in 3D. Our approach demonstrates greater view consistency, and is capable of generating high-quality 3D shapes. Furthermore, our experiments report superior FID image scores on FFHQ and AFHQ.

\section{Tri-plane hybrid 3D representation}
\reducespace
\label{sec:3d_representation_and_rendering}
Training a high-resolution GAN requires a 3D representation that is both efficient and expressive. In this section, we introduce a new hybrid explicit--implicit tri-plane representation that offers both of these advantages. We introduce the representation in this section for a single-scene overfitting (SSO) experiment, before discussing how it is integrated in our GAN framework in the next section. 




In the tri-plane formulation, we align our explicit features along three axis-aligned orthogonal feature planes, each with a resolution of $N \times N \times C$ (Fig. \ref{fig:hybrid_representation}c) with $N$ being spatial resolution and $C$ the number of channels. We query any 3D position $x \in \mathbb{R}^3$ by projecting it onto each of the three feature planes, retrieving the corresponding feature vector (${F_{xy}}$, ${F_{xz}}$, ${F_{yz}}$) via bilinear interpolation, and aggregating the three feature vectors via summation. An additional lightweight decoder network, implemented as a small MLP, interprets the aggregated 3D features ${F}$ as color and density. These quantities are rendered into RGB images using (neural) volume rendering~\cite{Max:1995,mildenhall2020nerf}. 

\begin{figure}[t!]
	\centering
		\includegraphics[width=0.8\columnwidth]{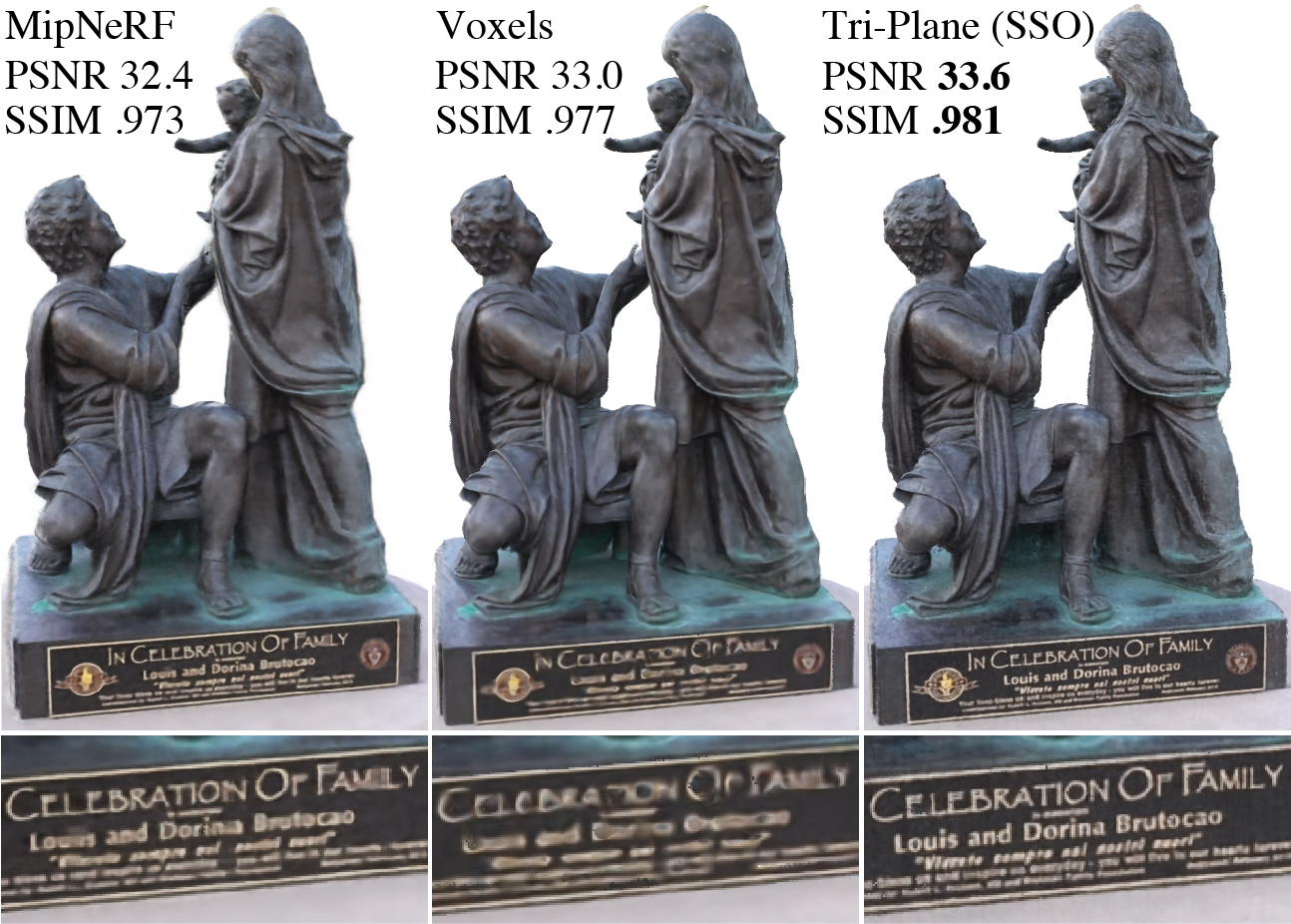}
		\caption{
		A synthesized view of the multi-view \textit{Family} scene, comparing a fully implicit Mip-NeRF representation (left), a dense voxel grid (center), and our tri-plane representation (right). Even though neither voxels nor tri-planes model view-dependent effects, they achieve high quality.
		}
		\label{fig:overfitting}
\end{figure}

\begin{table}
\centering
		\small
    \begin{tabular}{l c c c}
\toprule
               & MLP & Rel. Speed $\uparrow$ & Rel. Mem. $\downarrow$   \\ 
        \midrule
Mip-NeRF~\cite{barron2021mipnerf} &$8\times256$      & $1\times$         & $1\times$ \\
Voxels (hybrid) &$4\times128$   & $3.5\times$       & $0.33\times$ \\
Tri-plane (SSO) &$4\times128$ & $2.9\times$       & $0.32\times$  \\
Tri-plane (GAN) &$1\times64$  & $7.8\times$       & $0.06\times$  \\
\bottomrule
\end{tabular}
\caption{Relative speedups and memory consumption compared to Mip-NeRF. The proposed tri-plane representation is 3--8$\times$ faster than a fully implicit Mip-NeRF network and only requires a fraction of its memory. In this example, both voxel grid and tri-plane representation use an MLP-based decoder, as indicated. The number of voxels is chosen to match the total parameters of the tri-plane representation, thus the resolution is relatively low and the memory footprint lower than Mip-NeRF. In the SSO experiment (Fig.~\ref{fig:overfitting}), we used a larger decoder for the tri-plane representation than for the GAN experiments discussed in Sec.~\ref{sec:gan_framework} to optimize expressiveness over speed for this experiment.
}
\label{tab:overfitting_speed}
\end{table}


\begin{figure*}[t]
	\footnotesize
	\centering
	\includegraphics[width=\linewidth]{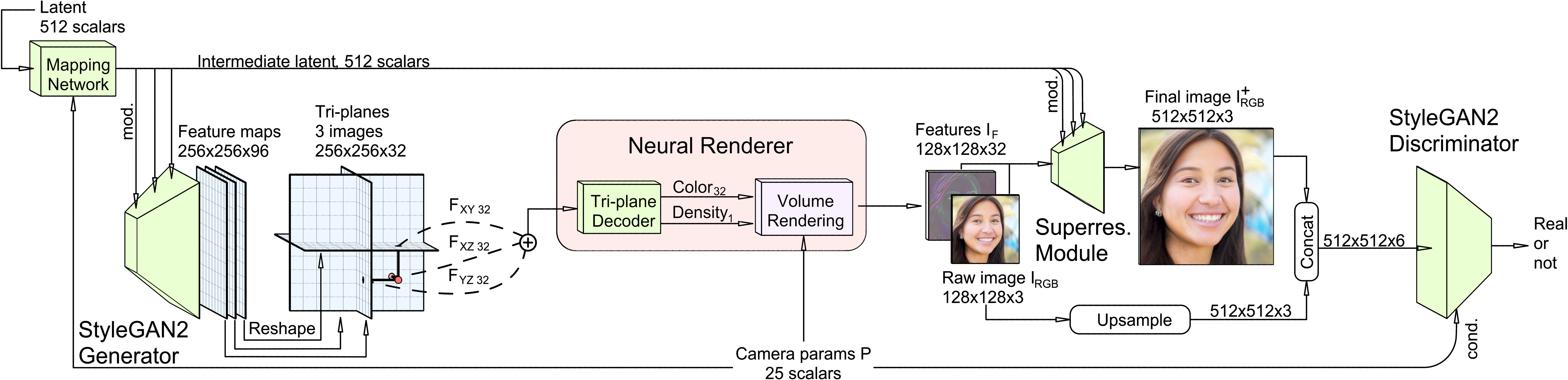}
	\caption{Our 3D GAN framework comprises several parts: a pose-conditioned StyleGAN2-based feature generator and mapping network, a tri-plane 3D representation with a lightweight feature decoder, a neural volume renderer, a super-resolution module, and a pose-conditioned StyleGAN2 discriminator with dual discrimination. 
	This architecture elegantly decouples feature generation and neural rendering, allowing the use of a powerful StyleGAN2 generator for 3D scene generalization. Moreover, the lightweight 3D tri-plane representation is both expressive and efficient in enabling high-quality 3D-aware view synthesis in real-time. 
	}
	\label{fig:pipeline}
	\vspace{-15pt}
\end{figure*}

The primary advantage of this hybrid representation is efficiency---by keeping the decoder small and shifting the bulk of the expressive power into the explicit features, we reduce the computational cost of neural rendering compared to fully implicit MLP architectures~\cite{mildenhall2020nerf,barron2021mipnerf} without losing expressiveness. To validate that the tri-plane representation is compact yet sufficiently expressive, we evaluate it with a common novel-view synthesis setup. For this purpose, we directly optimize the features of the planes and the weights of the decoder to fit $360^\circ$ views of a scene from the Tanks \& Temples dataset~\cite{Knapitsch2017} (Fig.~\ref{fig:overfitting}). In this experiment, we use feature planes of resolution $N=512$ and channels $C=48$, paired with an MLP of four layers of 128 hidden units each and a Fourier feature encoding\cite{tancik2020fourier}. We compare the results against a dense feature volume of equal capacity.
For reference, we include comparisons to a state-of-the-art fully implicit 3D representation~\cite{barron2021mipnerf}. Fig.~\ref{fig:overfitting} and Tab.~\ref{tab:overfitting_speed} demonstrate that the tri-plane representation is capable of representing this complex scene, albeit without view-dependent effects, outperforming dense feature volume representations~\cite{sitzmann2019deepvoxels,Lombardi:2019} and fully implicit representations~\cite{mildenhall2020nerf} in terms of PSNR and SSIM, while offering considerable advantages in computation and memory efficiency.
For a side length of $N$ features, tri-planes scale with $O(N^2)$ rather than $O(N^3)$ as dense voxels do, which means for equal capacity and memory, the tri-plane representation can use higher resolution features and capture greater detail. 
%
Finally, our tri-plane representation has one other key advantage over these alternatives: the feature planes can be generated with an off-the-shelf 2D CNN-based generator, enabling generalization across 3D representations using the GAN framework discussed next. 




\section{3D GAN framework}
\reducespace
\label{sec:gan_framework}


Armed with an efficient and expressive 3D representation, we train a 3D GAN for geometry-aware image synthesis from 2D photographs, without any explicit 3D or multi-view supervision. We associate each training image with a set of camera intrinsics and extrinsics using off-the-shelf pose detectors~\cite{deng2019accurate, cat_hipster}; see the supplement for details.
%

Fig.~\ref{fig:pipeline} gives an overview of our network architecture. We use the tri-plane representation introduced in the last section to efficiently render images through neural volume rendering, but make a number of modifications to adapt this representation to the 3D GAN setting. Unlike in the SSO experiment, where the features of the planes were directly optimized from the multiple input views, for the GAN setting we generate the tri-plane features, each containing $32$ channels, with the help of a 2D convolutional StyleGAN2 backbone (Sec.~\ref{sec:backbone_rendering}). Instead of producing an RGB image, in the GAN setting our neural renderer aggregates features from each of the $32$-channel tri-planes and predicts 32-channel feature images from a given camera pose. This is followed by a ``super-resolution" module to upsample and refine these raw neurally rendered images (Sec.~\ref{sec:super-resolution}). The generated images are critiqued by a slightly modified StyleGAN2 discriminator (Sec.~\ref{sec:dual-discrimination}). The entire pipeline is trained end-to-end from random initialization, using the non-saturating GAN loss function~\cite{goodfellow2014generative} with R1 regularization~\cite{gan_convergence}, following the training scheme in StyleGAN2~\cite{Karras2020stylegan2}.
To speed training, we use a two-stage training strategy in which we train with a reduced ($64^2$) neural rendering resolution followed by a short fine-tuning period at full ($128^2$) neural rendering resolution.
Additional experiments found that regularization to encourage smoothness of the density field helped reduce artifacts in 3D shapes.
The following sections discuss major components of our framework in detail. For additional descriptions, implementation details, and hyperparameters, please see the supplement. 

\subsection{CNN generator backbone and rendering}
\reducespace
\label{sec:backbone_rendering}

The features of the tri-plane representation, when used in our GAN setting, are generated by a StyleGAN2 CNN generator. The random latent code and camera parameters are first processed by a mapping network to yield an intermediate latent code which then modulates the convolution kernels of a separate synthesis network. 

%
We change the output shape of the StyleGAN2 backbone such that, rather than producing a three-channel RGB image, we produce a $256 \times 256 \times 96$ feature image.
This feature image is split channel-wise and reshaped to form three 32-channel planes (see Fig.~\ref{fig:pipeline}). We choose StyleGAN2 for predicting the tri-plane features because it is a well-understood and efficient architecture achieving state-of-the-art results for 2D image synthesis. Furthermore, our model inherits many of the desirable properties of StyleGAN: a well-behaved latent space that enables style-mixing and latent-space interpolation (see Sec.~\ref{sec:experiments} and supplement).


We sample features from the tri-planes, aggregate by summation, and process the aggregated features with a lightweight decoder, as described in Sec.~\ref{sec:3d_representation_and_rendering}. 
Our decoder is a multi-layer perceptron with a single hidden layer of $64$ units and softplus activation functions. The MLP does \textit{not} use a positional encoding, coordinate inputs, or view-direction inputs. 
This hybrid representation can be queried for continuous coordinates and outputs a scalar density $\sigma$ as well as a 32-channel feature, both of which are then processed by a neural volume renderer to project the 3D feature volume into a 2D feature image. 

%
Volume rendering~\cite{Max:1995} is implemented using two-pass importance sampling as in~\cite{mildenhall2020nerf}. Following~\cite{Niemeyer2020GIRAFFE}, volume rendering in our GAN framework produces feature images, rather than RGB images, because feature images contain more information that can be effectively utilized for the image-space refinement described next. For the majority of the experiments reported in this manuscript, we render 32-channel feature images $I_F$ at a resolution of $128^2$, with 96 total depth samples per ray.

\subsection{Super resolution}
\reducespace
\label{sec:super-resolution}

Although the tri-plane representation is significantly more computationally efficient than previous approaches, it is still too slow to natively train or render at high resolutions while maintaining interactive framerates. We thus perform volume rendering at a moderate resolution (e.g., $128^2$) and rely upon image-space convolutions to upsample the neural rendering to the final image size of $256^2$ or $512^2$.



Our super resolution module is composed of two blocks of StyleGAN2-modulated convolutional layers that upsample and refine the 32-channel feature image $I_{F}$ into the final RGB image $I^+_{RGB}$. We disable per-pixel noise inputs to reduce texture sticking~\cite{Karras2021} and reuse the mapping network of the backbone to modulate these layers.


\begin{figure}
    \centering
    \includegraphics[width=\linewidth]{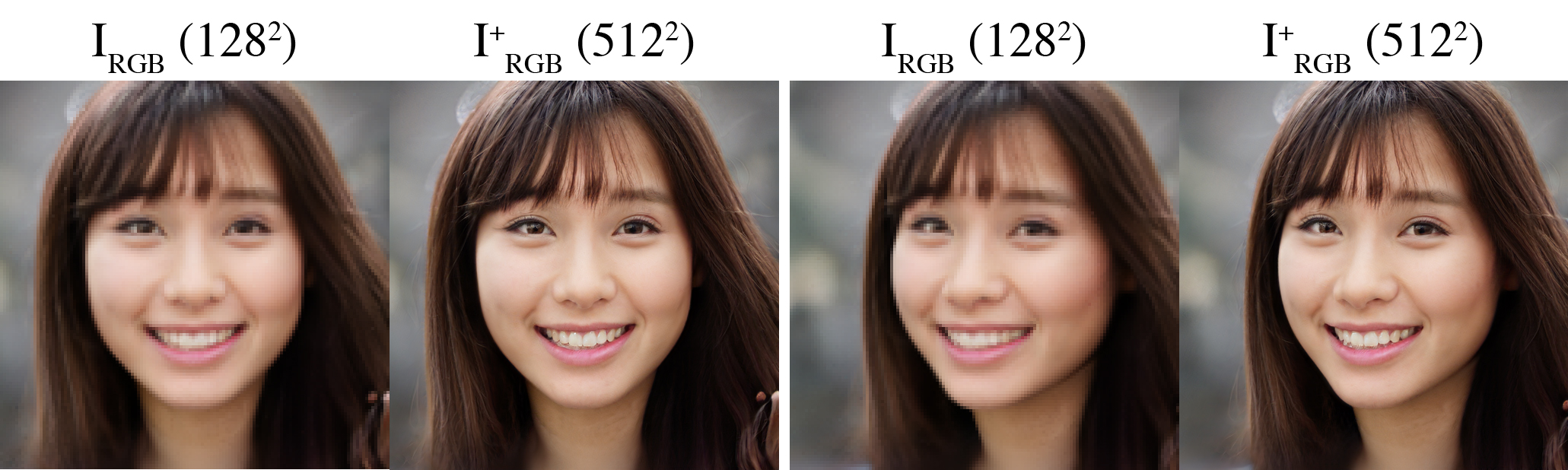}
    \caption{
    Dual discrimination ensures that the raw neural rendering $I_{RGB}$ and super-resolved output $I^+_{RGB}$ maintain consistency, enabling high-resolution and multi-view-consistent rendering.  
    }
    \label{fig:superresolution-dual-discrimination}
\end{figure}

\subsection{Dual discrimination}
\reducespace
\label{sec:dual-discrimination}

As in standard 2D GAN training, the resulting renderings are critiqued by a 2D convolutional discriminator. We use a StyleGAN2 discriminator with two modifications.

First, we introduce \textit{dual discrimination} as a method to avoid multi-view inconsistency issues observed in prior work~\cite{hologan,Niemeyer2020GIRAFFE}. For this purpose, we interpret the first three feature channels of a neurally rendered feature image $I_F$ as a low-resolution RGB image $I_{RGB}$. Intuitively, dual discrimination then ensures consistency between $I_{RGB}$ and the super-resolved image $I^+_{RGB}$. This is achieved by bilinearly upsampling $I_{RGB}$ to the same resolution as $I^+_{RGB}$ and concatenating the results to form a six-channel image (see Fig.~\ref{fig:pipeline}). The real images fed into the discriminator are also processed by concatenating each of them with an appropriately blurred copy of itself. We discriminate over these six-channel images instead of the three-channel images traditionally seen in GAN discriminators.

Dual discrimination not only encourages the final output to match the distribution of real images, but also offers additional effects: it encourages the neural rendering to match the distribution of downsampled real images; and it encourages the super-resolved images to be consistent with the neural rendering (see Fig.~\ref{fig:superresolution-dual-discrimination}). The second point importantly allows us to leverage effective image-space super-resolution layers without introducing view-inconsistency artifacts. 

Second, we make the discriminator aware of the camera poses from which the generated images are rendered. Specifically, following the conditional strategy from StyleGAN2-ADA~\cite{Karras2020ada}, we pass the rendering camera intrinsics and extrinsics matrices (collectively $\mathbf{P}$) to the discriminator as a conditioning label. 
%
We find that this conditioning introduces additional information that guides the generator to learn correct 3D priors.
We provide additional studies in the supplement showing the effect of this discriminator conditioning and the robustness of our framework to high levels of noise in the input camera poses.
\begin{figure*}
    \centering
    \includegraphics[width=\textwidth]{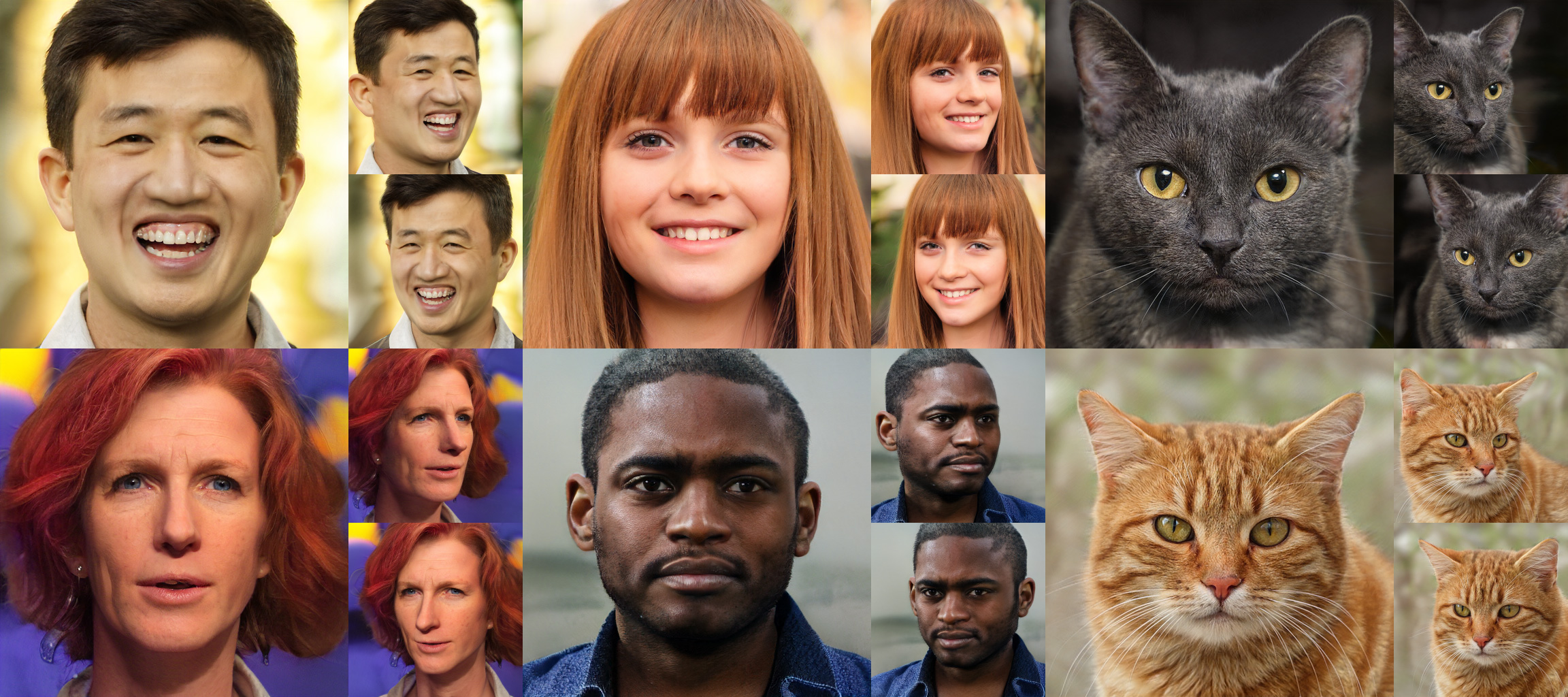}
    \caption{Curated examples at $512^2$, synthesized by models trained with FFHQ\cite{karras2019style} and AFHQv2 Cats\cite{choi2020starganv2}}
    \label{fig:curated_images}
\end{figure*}

\begin{figure*}
    \vspace{-3mm}
    \centering
    \includegraphics[width=\textwidth]{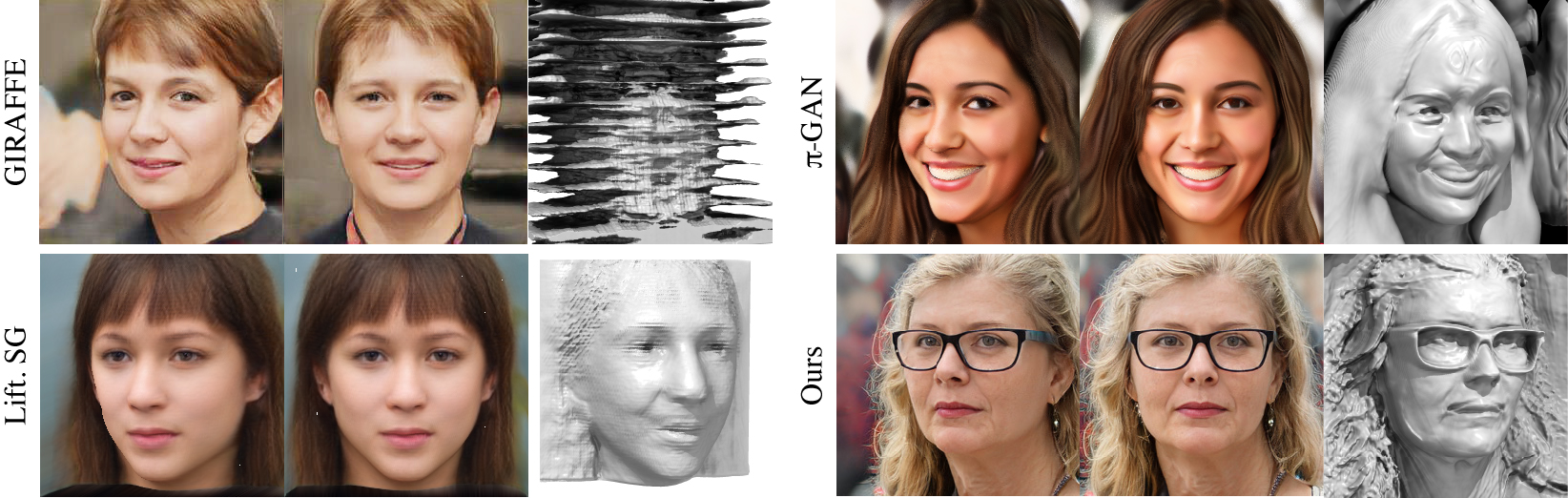}
    \caption{Qualitative comparison between GIRAFFE, $\pi$-GAN, Lifting StyleGAN, ours, with FFHQ at $256^2$. Shapes are iso-surfaces extracted from the density field using marching cubes. We inspected the underlying 3D representations of GIRAFFE and found that its over-reliance on image-space approximations significantly harms the learning of the 3D geometry.
    }
    \label{fig:qualitative-comparison}
    \vspace{-15pt}
\end{figure*}


\subsection{Modeling pose-correlated attributes}
\reducespace
\label{sec:generator_pose_conditioning}
Most real-world datasets like FFHQ include biases that correlate camera poses with other attributes (e.g., facial expressions), and naively handling them leads to view inconsistent results. For example, the camera angle with respect to a person's face is correlated with smiling (see supplement). 
While faithfully modeling such attribute correlations inherent in the dataset is important for reproducing the best image quality, such unwanted attributes need to be decoupled during inference for multi-view consistent synthesis. 
Related work has been successful at being view consistent~\cite{graf,chan2020pi,shi2021lifting} or modeling pose-appearance correlations~\cite{hologan,Niemeyer2020GIRAFFE}, but cannot achieve both simultaneously.

We introduce \textit{generator pose conditioning} as a means to model and decouple correlations between pose and other attributes observed in the training images. To this end, we provide the backbone mapping network not only a latent code vector $z$, but also the camera parameters $\mathbf{P}$
as input, following the conditional generation strategy in~\cite{Karras2020ada}. By giving the backbone knowledge of the rendering camera position, we allow the target view to influence scene synthesis. 

During training, pose conditioning allows the generator to model pose-dependent biases implicit to the dataset, allowing our model to faithfully reproduce the image distributions in the dataset. To prevent the scene from shifting with camera pose during inference, we condition the generator on a fixed camera pose when rendering from a moving camera trajectory. We noticed that always conditioning the generator with the rendering camera pose can lead to degenerate solutions where the GAN produces 2D billboards angled towards the camera (see supplement). 
To prevent this, we randomly swap the conditioning pose in $\mathbf{P}$ with another random pose with 50\% probability during training. 

\section{Experiments and results}
\reducespace
\label{sec:experiments}
\paragraph{Datasets.} We compare methods on the task of unconditional 3D-aware generation with FFHQ~\cite{karras2019style}, a real-world human face dataset, and AFHQv2 Cats~\cite{choi2020starganv2,Karras2021}, a small, real-world cat face dataset.
We augment both datasets with horizontal flips and use off-the-shelf pose estimators~\cite{deng2019accurate,cat_hipster} to extract approximate camera extrinsics. For all methods on AFHQv2, we apply transfer learning~\cite{Karras2020ada} from corresponding FFHQ checkpoints; for our method on AFHQv2 $512^2$, we additionally use adaptive data augmentation\cite{Karras2020ada}.
For more results, please see the accompanying video. 



\subsection{Comparisons}
\reducespace
\paragraph{Baselines.} We compare our methods against three state-of-the-art methods for 3D-aware image synthesis: $\pi$-GAN~\cite{chan2020pi}, GIRAFFE~\cite{Niemeyer2020GIRAFFE}, and Lifting StyleGAN~\cite{shi2021lifting}.

\myparagraph{Qualitative results.}
Fig.~\ref{fig:curated_images} presents selected examples synthesized by our model with FFHQ and AFHQ at a resolution of $512^2$, highlighting the image quality, view-consistency, and diversity of outputs produced by our method. Fig.~\ref{fig:qualitative-comparison} provides a qualitative comparison against baselines.
While GIRAFFE synthesizes high-quality images, reliance on view-inconsistent convolutions produces poor-quality shapes and identity shift---note the hairline inconsistency between rendered views.
$\pi$-GAN and Lifting StyleGAN generate adequate shapes and images but both struggle with photorealism and in capturing detailed shapes.


Our method synthesizes not only images that are higher quality and more view-consistent but also higher-fidelity 3D geometry as seen in the detailed glasses and hair strands. 




\begin{table}[t]
    \centering
		\small
    \begin{tabular}{@{\hskip 1mm}l c c c c c@{\hskip 1mm}}
				\toprule
					& \multicolumn{4}{c}{FFHQ} & Cats\\
					            & FID $\!\downarrow$        & ID $\!\uparrow$   & Depth $\!\downarrow$  & Pose $\!\downarrow$   & FID $\!\downarrow$  \\
        \midrule
        GIRAFFE $256^2$	            & 31.5 					    & 0.64 	            & 0.94 					& .089 				    & 16.1                  \\
        $\pi$-GAN $128^2$ 	& 29.9 				    	& 0.67 	            & 0.44 					& .021 				    & 16.0                \\
        Lift. SG $256^2$	            & 29.8				    	& 0.58              & 0.40					& .023				                      & --- \\
        Ours $256^2$ 			& \textbf{4.8}              & 0.76 	            & \textbf{0.31}	        & \textbf{.005}		    & \textbf{3.88}        \\
        Ours $512^2$ 			& \textbf{4.7}              & \textbf{0.77} 	& 0.39	                & \textbf{.005}         & $\textbf{2.77}^\dag$      \\
				\bottomrule
    \end{tabular}
    \caption{Quantitative evaluation using FID, identity consistency (ID), depth accuracy, and pose accuracy for FFHQ and AFHQ Cats. Labelled is the image resolution of training and evaluation. \textsuperscript{\dag} Trained with adaptive data augmentation~\cite{Karras2020ada}.
    }
    \label{tab:main_results}
\end{table}


\myparagraph{Quantitative evaluations.}
Table~\ref{tab:main_results} provides quantitative metrics comparing the proposed approach against baselines. We measure image quality with Fr\'echet Inception Distance ({FID})~\cite{DBLP:journals/corr/HeuselRUNKH17} between 50k generated images and all available real images.  
We evaluate shape quality by calculating MSE against pseudo-ground-truth depth-maps ({Depth}) and poses ({Pose}) estimated from synthesized images by~\cite{deng2019accurate}; a similar evaluation was introduced by \cite{shi2021lifting}.
We assess multi-view facial identity consistency ({ID}) by calculating the mean Arcface\cite{deng2018arcface} cosine similarity score between pairs of views of the same synthesized face rendered from random camera poses. Additional evaluation details are provided in the supplement. Our model demonstrates significant improvements in FID across both datasets, bringing the 3D GAN to near the same level as StyleGAN2 $512^2$ (2.97 for FFHQ~\cite{Karras2020stylegan2} and 2.99 for Cats~\cite{Karras2020ada}) while also maintaining state-of-the-art view consistency, geometry quality, and pose accuracy.


\begin{table}[t]
    \centering
		\small
    \begin{tabular}{@{\hskip 1mm}l c c c c c@{\hskip 1mm}}
		\toprule
		Res. & GIRAFFE & $\pi$-GAN & Lift. SG & Ours & Ours + TC \\
		\midrule
		$256^2$ & 181 & 5 & 51 & 27 & 36 \\
		$512^2$ & 161 & 1 & --- & 26 & 35 \\
		\bottomrule
    \end{tabular}
    \caption{Runtime in frames per second at different rendering resolutions. We compare variants of our approach with and without tri-plane caching (TC). Run on a single RTX 3090 GPU.}
    \label{tab:runtime}
\end{table}


\myparagraph{Runtime.}
Table~\ref{tab:runtime} compares rendering speed at inference running on a single NVIDIA RTX 3090 GPU. Our end-to-end approach achieves real-time framerates at $512^2$ final resolution with $128^2$ neural rendering resolution and 96 total depth samples per ray, suitable for applications such as real-time visualization. 
When rendering consecutive frames of a static scene, we need not regenerate the tri-plane features every frame; caching the generated features is a simple tweak that improves render speed.
The proposed approach is significantly faster than fully implicit methods like $\pi$-GAN~\cite{chan2020pi}. Although it is not as fast as Lifting StyleGAN~\cite{shi2021lifting} and GIRAFFE~\cite{Niemeyer2020GIRAFFE}, we believe major improvements in image quality, geometry quality, and view-consistency outweigh the increased compute cost.

\subsection{Ablation study}
\reducespace


\begin{table}[t]
    \centering
		\small
    \begin{tabular}{l c c}
				\toprule
										& FID $\downarrow$ & FACS Smile Std. $\downarrow$ \\
				\midrule
				Naive model 							& 5.5 & 0.069 \\
        + DD & 6.5 & 0.054 \\
				+ DD, GPC (ours) 						& \textbf{4.7} & \textbf{0.031} \\				
				\bottomrule
    \end{tabular}
    \caption{Dual-discrimination (DD) improves multi-view expression consistency but hurts the model's ability to capture pose-correlated attributes for image quality. Adding generator pose conditioning (GPC) allows the model to improve upon both aspects. Reported at $512^2$, with FFHQ.}
    \label{tab:ablation}
\end{table}


Without dual discrimination, generated images can include multi-view inconsistencies due to the unconstrained image-space super-resolution layers. We measure this effect quantitatively by extracting smile-related Facial Action Coding System (FACS)~\cite{Ekman1978} coefficients from videos produced by models with and without dual discrimination, using a proprietary facial tracker.
We measure the standard deviation of smile coefficients for the same scene across video frames. A view-consistent scene should exhibit little expression shift and thus produce little variation in smile coefficients. This is validated in Table \ref{tab:ablation} showing that introducing dual discrimination (second row) reduces the smile coefficient variation versus the naive model (first row), indicating improved expression consistency. However, dual discrimination also reduces image quality as seen by the slightly worse FID score, perhaps because the model is restricted from reproducing the pose-correlated attribute biases in the FFHQ dataset. 
By adding generator pose conditioning (third row), we allow the generator to faithfully model pose-correlated attributes while decoupling them at inference, leading to both the best FID score and view-consistent results. 
\subsection{Applications}

\begin{figure}[t]
    \centering
    \includegraphics[width=\linewidth]{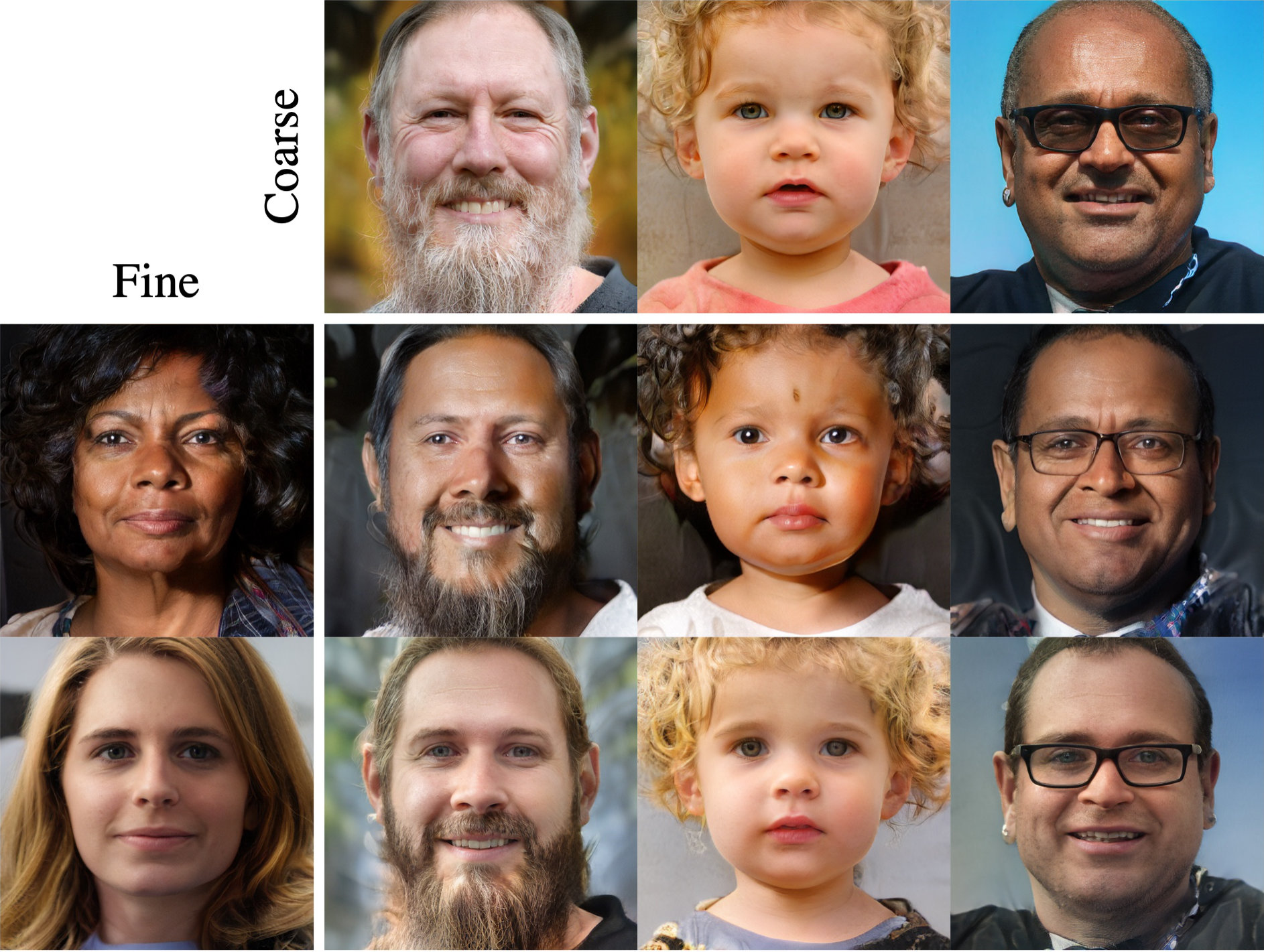}
    \caption{Style-mixing\cite{karras2019style,Karras2020stylegan2,Karras2021} with FFHQ $512^2$.
    } 
    \label{fig:style_mix}
\end{figure}

\paragraph{Style mixing.}
Since our 3D representation is designed with the StyleGAN2 backbone from the ground up, it inherits the well-studied properties of the StyleGAN2 latent space, allowing us to do semantic image manipulations. Fig.~\ref{fig:style_mix} shows our method's results for style mixing~\cite{karras2019style,Karras2020stylegan2,Karras2021}. 

\begin{figure}
    \centering
    \includegraphics[width=\linewidth, keepaspectratio=true]{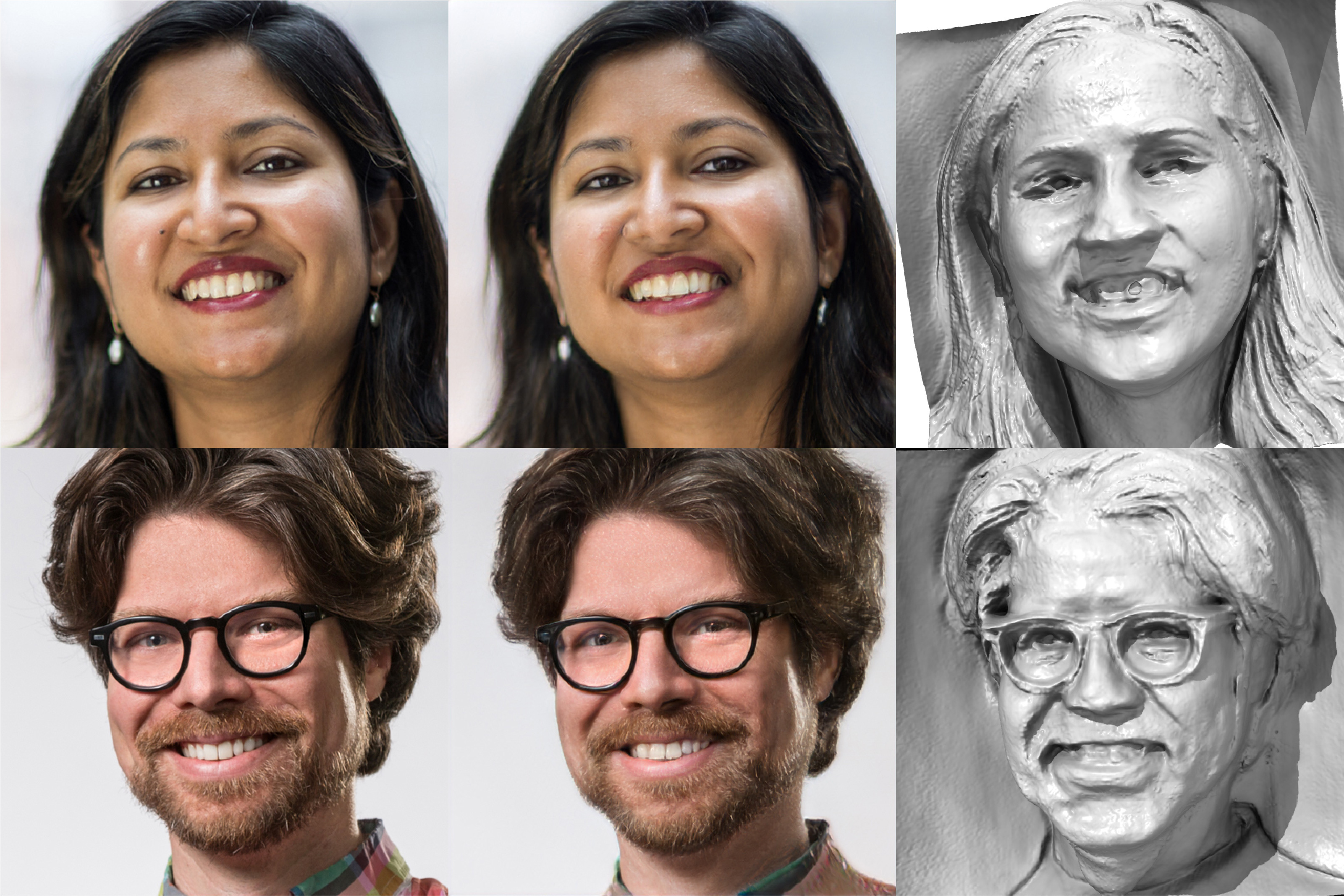}
    \caption{
    We use PTI\cite{roich2021pivotal} to fit a target image and recover the underlying 3D shape. Target (left); reconstructed image (center); reconstructed shape (right).
    From a model trained on FFHQ $512^2$.
    }
    \label{fig:inversion}
\end{figure}

\myparagraph{Single-view 3D reconstruction.} 
Fig.~\ref{fig:inversion} shows the application of our learned latent space for single-view 3D reconstruction. We use pivotal tuning inversion (PTI)~\cite{roich2021pivotal} to fit test images. The learned 3D prior over FFHQ enables surprisingly high-quality single-view geometry recovery. Further exploration of few-shot 3D reconstruction and novel-view-synthesis may prove a fruitful avenue for future work.

\section{Discussion}
\vspace{-5pt}
\label{sec:discussion}
\paragraph{Limitations and future work.}
Although our shapes show significant improvements over those generated by previous 3D-aware GANs, they may still contain artifacts and lack finer details, such as individual teeth. To further improve the quality of the learned shapes, we could instill a stronger geometry prior or regularize the density component of the radiance field following methods proposed by~\cite{yariv2021volume, wang2021neus, oechsle2021unisurf}.


%
Our model requires knowledge of the camera pose distribution of the dataset. Although prior work has proposed learning the pose distribution on the fly~\cite{Niemeyer2020GIRAFFE}, others have noticed such methods can diverge~\cite{gu2021stylenerf}, so it would be fruitful to explore this direction further. Pose conditioning aids the generator in decoupling appearance from pose, but still does not fully disentangle the two.
Furthermore, ambiguities that can be explained by geometry remain unresolved. For example, by creating concave eye sockets, the generator creates the illusion of eyes that ``follow'' the camera, an incorrect interpretation, though the renderings are view-consistent and reflect the underlying geometry.


We used StyleGAN 2, but other 2D backbones may find success in our framework. Alternative backbones, such as as image-to-image translation or Transformer-based models, could enable new applications in conditional synthesis.


\myparagraph{Ethical considerations.}
The single-view 3D reconstruction or style mixing applications could be misused for generating edited imagery of real people. Such misuse of image synthesis techniques poses a societal threat, and we do not condone using our work with the intent of spreading misinformation or tarnishing reputation. We also recognize a potential lack of diversity in our faces results, stemming from implicit biases of the datasets we process.

\myparagraph{Conclusion.}
By combining an efficient explicit--implicit neural representation with an expressive pose-aware convolutional generator and a dual discriminator, our approach takes significant steps towards photorealistic 3D-aware image synthesis and high-quality unsupervised shape generation. This may enable rapid prototyping of 3D models, more controllable image synthesis, and novel techniques for shape reconstruction from temporal data.

\section*{Acknowledgements}
\vspace{-5pt}
We thank David Luebke, Jan Kautz, Jaewoo Seo, Jonathan Granskog, Simon Yuen, Alex Evans, Stan Birchfield, Alexander Bergman, and Joy Hsu for feedback on drafts, Alex Chan, Giap Nguyen, and Trevor Chan for help with diagrams, and Colette Kress and Bryan Catanzaro for allowing use of their photographs. This project was in part supported by Stanford HAI and a Samsung GRO. Koki Nagano and Eric Chan were partially supported by DARPA’s Semantic Forensics (SemaFor) contract (HR0011-20-3-0005). The views and conclusions contained in this document are those of the authors and should not be interpreted as representing the official policies, either expressed or implied, of the U.S. Government. Distribution Statement ``A'' (Approved for Public Release, Distribution Unlimited).

{\small
\bibliographystyle{ieee_fullname}
\bibliography{egbib}
}


\end{document}


\title{Supplemental Material\\Efficient Geometry-aware 3D Generative Adversarial Networks}

\author{First Author\\
Institution1\\
Institution1 address\\
{\tt\small firstauthor@i1.org}
\and
Second Author\\
Institution2\\
First line of institution2 address\\
{\tt\small secondauthor@i2.org}
}
\maketitle
 
\global\csname @topnum\endcsname 0
\global\csname @botnum\endcsname 0



In this supplement, we first provide additional experiments (Section~\ref{sec:supp_additional_experiments}) and visual results (Section~\ref{sec:supp_additional_visual_results}). We follow with details of our implementation (Section~\ref{sec:supp_implementation_details}), including further descriptions of model architecture and training process, as well as hyperparameters. We discuss experiment details (Section~\ref{sec:supp_experiment_details}), such as datasets and baselines, and further explanations for experiments such as inversion. Lastly, we consider artifacts (Section~\ref{sec:supp_discussion}) that may be targets of future work. We encourage readers to view the accompanying supplemental video, which contains additional visual results, including a live demonstration of real-time synthesis.

\section{Additional experiments}
\label{sec:supp_additional_experiments}

\subsection{Analyzing pose/facial expression correlation in FFHQ}

\begin{figure}[h!]
    \centering
    \includegraphics[width=\linewidth, keepaspectratio=true]{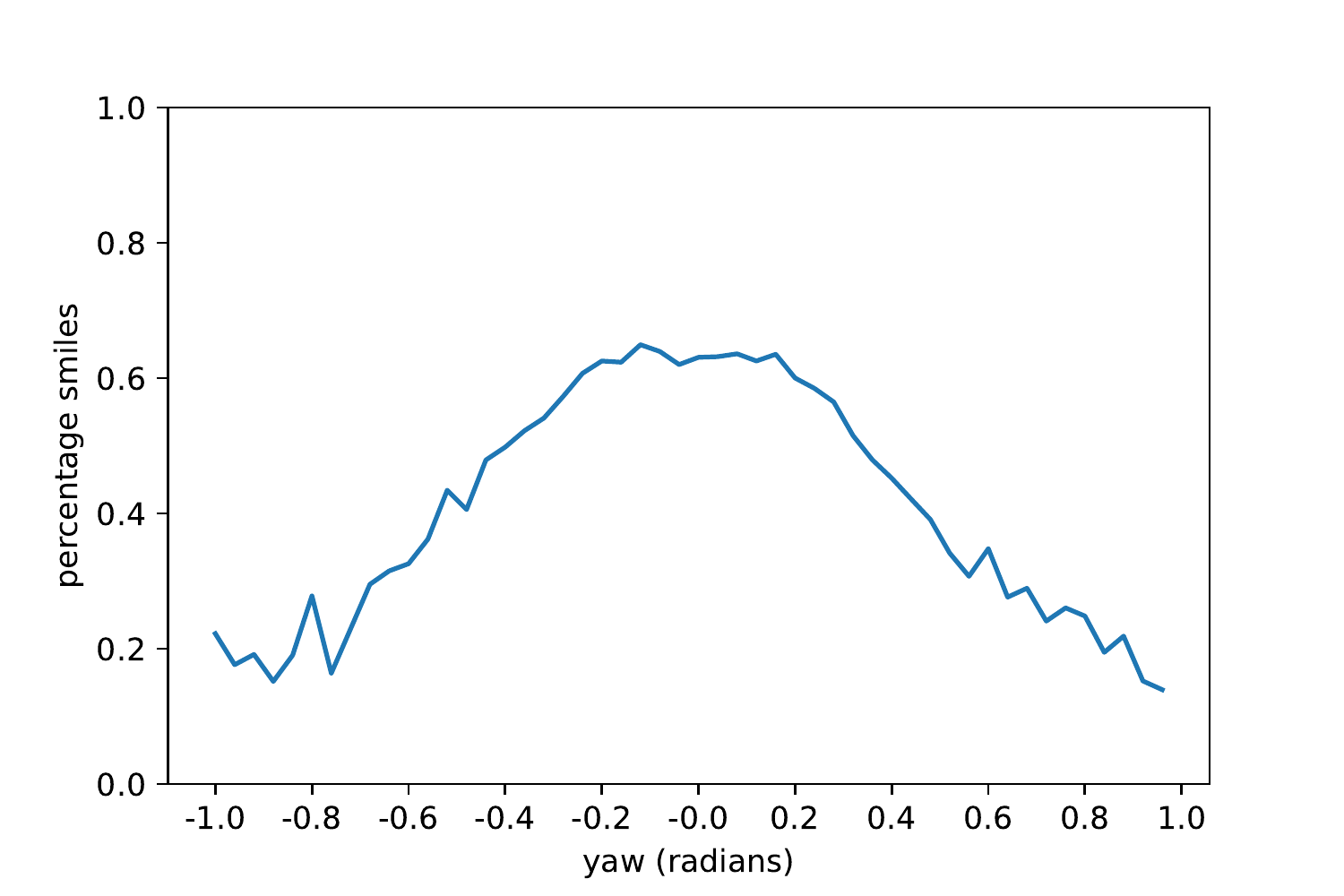}
    \caption{We plot the probability of smiling against head yaw angle, as measured by \cite{expression_detection}. People looking at the camera are more likely to be smiling than people angled away, indicating a correlation between scene appearance and camera pose.}
    \label{fig:smiliness}
\end{figure}

Fig.~\ref{fig:smiliness} plots the likelihood a subject from FFHQ \cite{karras2019style} is smiling (measured by \cite{expression_detection}), against head yaw (computed by \cite{deng2019accurate}). The plot indicates that individuals facing towards the camera are more likely to be smiling than are individuals who are facing away from the camera. An intuitive explanation for this phenomenon is that people who are knowingly being photographed, as in portrait images, are more likely to be smiling than people who are photographed candidly.

Left uncompensated for, this correlation between pose and facial expressions incentivizes ``expression warping'', where the expressions of synthesized faces shift as we move the camera. We propose dual discrimination (Section 4.3 of the main paper) and generator pose conditioning (Section 4.4 of the main paper) to reduce such expression warping.

\subsection{COLMAP reconstruction}

\begin{figure}[h!]
    \centering
    \includegraphics[width=\linewidth]{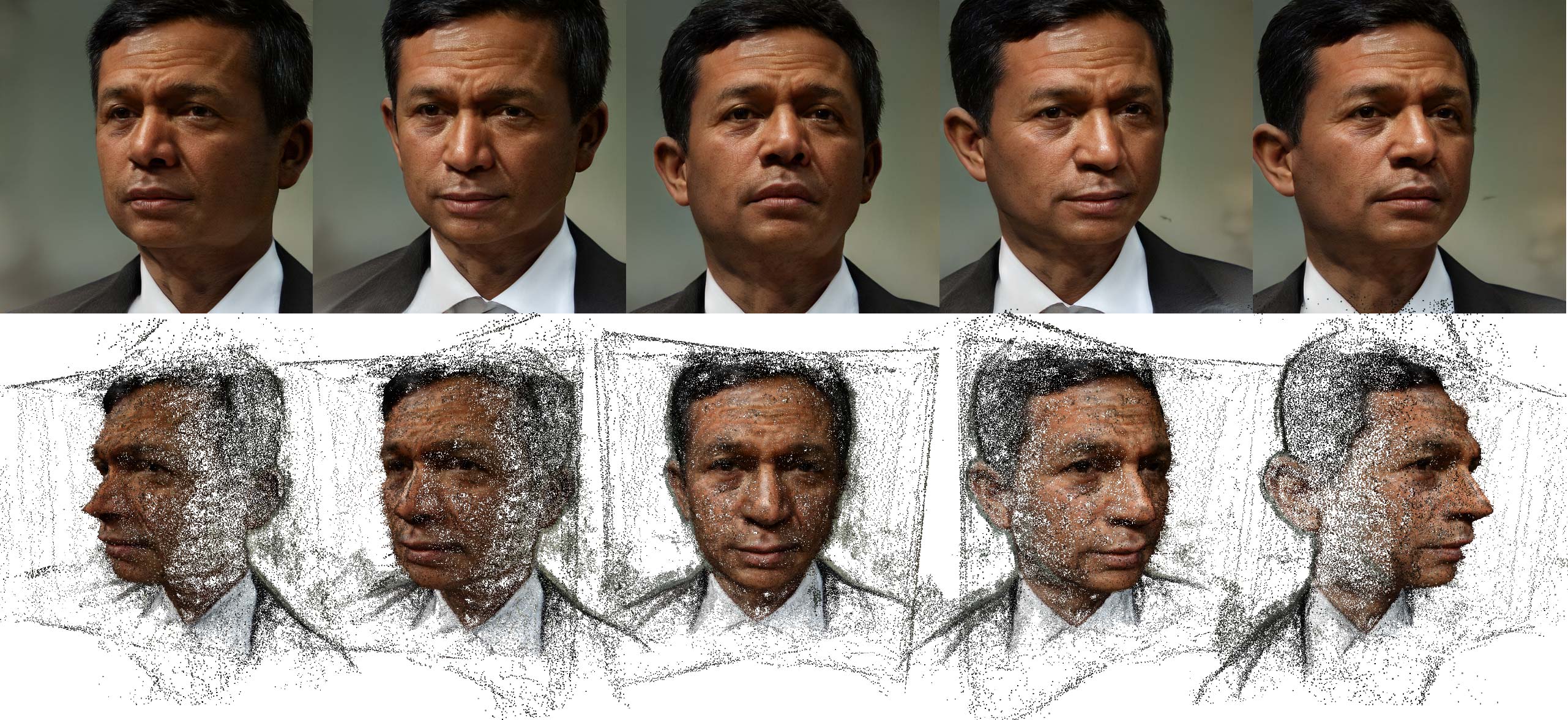}
    \caption{
    COLMAP\cite{schoenberger2016sfm, schoenberger2016mvs} reconstruction of 128 frames of synthesized video (top) which followed an oval trajectory. The resulting dense, well-defined point cloud (bottom) is indicative of highly multi-view-consistent rendering.}
    \label{fig:colmap}
\end{figure}


To further validate the multi-view consistency of our method, we employ COLMAP\cite{schoenberger2016sfm, schoenberger2016mvs} to reconstruct a point-cloud of a synthesized video sequence (Fig.~\ref{fig:colmap}). We reconstruct a video sequence of 128 frames, taken from an oval trajectory similar to the camera paths shown in the supplemental video. We use COLMAP's ``automatic'' reconstruction, without specifying camera parameters. The resulting point cloud is dense and well-defined, indicating that our 3D GAN produces highly multi-view-consistent renderings.

\subsection{Regularizing generator pose conditioning}
\begin{figure}[h!]
    \centering
    \includegraphics[width=\linewidth]{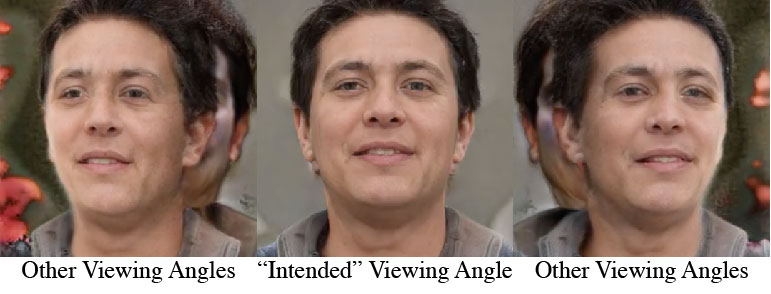}
    \caption{Naively applying generator pose conditioning results in a degenerate solution because the generator is always aware of the location of the rendering camera. Such an approach produces reasonable renderings when taken from the ``intended'' viewing angle, (i.e. the camera pose the generator was conditioned on). However, if we freeze the conditioning information and move the camera at inference, it is clear that the model has learned to produce ``billboards'' angled towards the known location of the camera.}
    \label{fig:naive-gpc}
\end{figure}


As described in Section 4.4 of the main paper, we regularize generator pose conditioning by randomly swapping the conditioning pose of the generator with another random pose with 50\% probability. Fig.~\ref{fig:naive-gpc} shows the result of training a model with generator pose conditioning but without any swapping regularization—the generator always receives, as a conditioning input, the true pose of the rendering camera. The model learns a degenerate solution in which it creates a ``billboard'' angled towards the rendering camera. We prevent this degenerate solution by randomly swapping the conditioning camera pose with an alternative pose sampled from the dataset pose distribution. For models shown, we swap the conditioning vector with 100\% probability at the start of training; the swapping probability is linearly decayed to 50\% over the first 1M images. For the remainder of training, we maintain 50\% swapping probability.

\begin{table*}[t]
    \centering
		\small
    \begin{tabular}{@{\hskip 1mm}l c c c c c c c c c@{\hskip 1mm}}
				\toprule
					       & \multicolumn{5}{c}{FFHQ}                                                                                           & \multicolumn{2}{c}{Cats}                & \multicolumn{2}{c}{Cars}\\
					       & FID $\!\downarrow$     & KID $\!\downarrow$   & ID $\!\uparrow$   & Depth $\!\downarrow$  & Pose $\!\downarrow$    & FID $\!\downarrow$ & KID $\!\downarrow$ & FID $\!\downarrow$ & KID $\!\downarrow$\\
        \midrule
        GIRAFFE $128^2$	    & ---           &	---          & ---           & --- 	          & ---	          & ---                  & ---                   & 27.3          & 1.703 \\
        GIRAFFE $256^2$	    & 31.5          &	1.992        & 0.64          & 0.94 	      & .089 	      & 16.1                 & 2.723                 & ---           & --- \\
        $\pi$-GAN $128^2$ 	& 29.9          &	3.573 	     & 0.67 	     & 0.44 		  & .021 		  & 16.0                 & 1.492                 & 17.3          & 0.932  \\
        Lift. SG $256^2$    & 29.8	        &	---          & 0.58          & 0.40		      & .023          & ---                  & ---                   & ---           & --- \\
        
        Ours $128^2$ 		& ---           & ---            & --- 	         & ---	          & ---		      & ---                  & ---                   & \textbf{2.75} & \textbf{0.097} \\
        Ours $256^2$ 		& \textbf{4.8}  & \textbf{0.149} & 0.76 	     & \textbf{0.31}  & \textbf{.005} & \textbf{3.88}        & \textbf{0.091}        & ---           & --- \\
        Ours $512^2$ 		& \textbf{4.7}  & \textbf{0.132} & \textbf{0.77} & 0.39	          & \textbf{.005} & $\textbf{2.77}^\dag$ & $\textbf{0.041}^\dag$ & ---           & --- \\
				\bottomrule
    \end{tabular}
    \captionsetup{width=.73\textwidth}
    \caption{Quantitative evaluation using FID, KID$\times 100$, identity consistency (ID), depth accuracy, and pose accuracy for FFHQ~\cite{karras2019style} and FID, KID$\times 100$ for AFHQv2 Cats~\cite{choi2020starganv2, Karras2021} and ShapeNet Cars~\cite{chang2015shapenet, sitzmann2019srns}. Labeled is the image resolution of training and evaluation. \textsuperscript{\dag} Trained with adaptive discriminator augmentation~\cite{Karras2020ada}.
    }
    \label{tab:supp_main_results}
\end{table*}

\subsection{Robustness to imprecise camera poses}

\begin{figure}
    \centering
    \includegraphics[width=\linewidth]{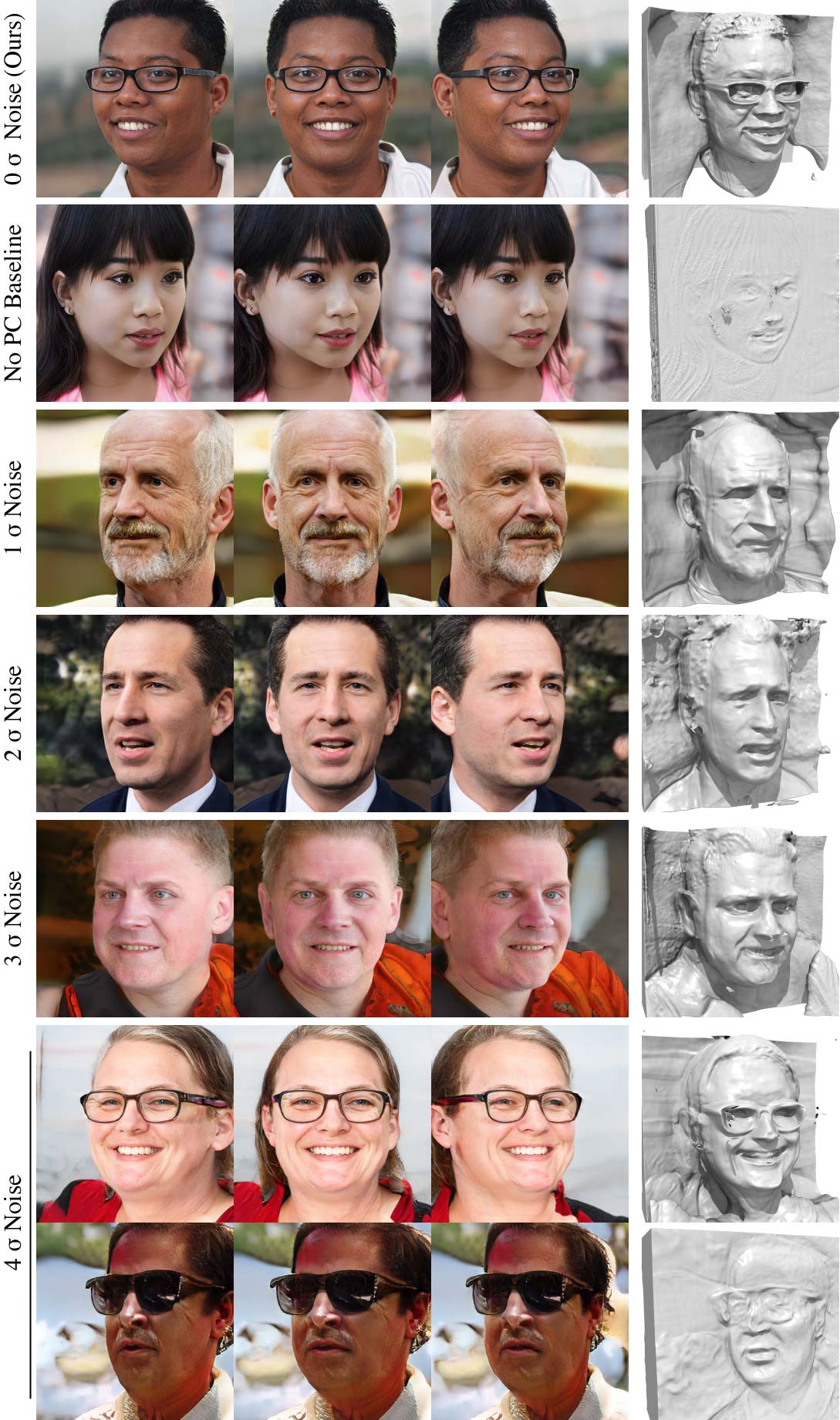}
    \caption{In order to gauge robustness to the accuracy of the supplied camera poses, we compare a baseline without discriminator pose conditioning against discriminator-pose-conditioned models where camera extrinsics are corrupted by noise. Without discriminator pose conditioning, the model learns a degenerate solution in which heads are drawn as a texture flattened onto a plane. Even highly imprecise extrinsics (e.g. camera poses corrupted by three standard deviations of noise) are capable of resolving this degenerate solution and allow recovery of accurate 3D shapes.}
    \label{fig:pose-noise-ablation}
\end{figure}

Our method expects a dataset in which each image is labeled with an approximate camera pose, in order to enable sampling camera poses from the dataset distribution and discriminator pose conditioning. While such labelling can be easily performed with pre-trained pose extractors on humans~\cite{deng2019accurate} and cats~\cite{cat_hipster}, extracting accurate poses may be difficult for some datasets. This section evaluates reliance on discriminator pose conditioning and on accurate camera poses. We train five additional models on FFHQ $256^2$: a ``baseline'' configuration without discriminator pose conditioning, and four discriminator-pose-conditioned models where camera poses are corrupted with increasing levels of random noise. We calculate the $4 \times 4$ standard deviation matrix, $\sigma$, by taking the standard deviation across the dataset of ground-truth $4 \times 4$ camera pose matrices. We train four models with ``imprecise'' camera poses: (1 $\sigma$, 2 $\sigma$, 3 $\sigma$, 4 $\sigma$) where the input camera poses matrices are corrupted with 1, 2, 3, and 4 standard deviations of Gaussian noise, respectively. We train these five ablations on FFHQ $256^2$ with a shortened training curriculum of 4M images, in order to save computational resources. 

Fig.~\ref{fig:pose-noise-ablation} shows the results of this experiment. Without discriminator pose conditioning, the model falls into a degenerate solution in which it renders textures on a flat plane, without properly capturing the 3D shape of scenes.
Providing even very imprecise camera poses is enough to break this tendency; conditioning the discriminator on camera poses distorted by three standard deviations of Gaussian noise still produces accurate 3D shapes. With extreme noise (e.g. four standard deviations), some scenes maintain the correct 3D structure while others are flattened onto the plane. Our results indicate that while our method requires additional information to prevent collapse, only very weak supervision is necessary. Future work may examine this tendency further and discover ways to prevent this undesirable behavior without requiring images to be labelled with poses.




\subsection{Extrapolation to steep camera angles}

\begin{figure*}
    \centering
    \captionsetup{width=.67\linewidth}
    \begin{subfigure}[b]{0.67\textwidth}
         \centering
         \includegraphics[width=\textwidth]{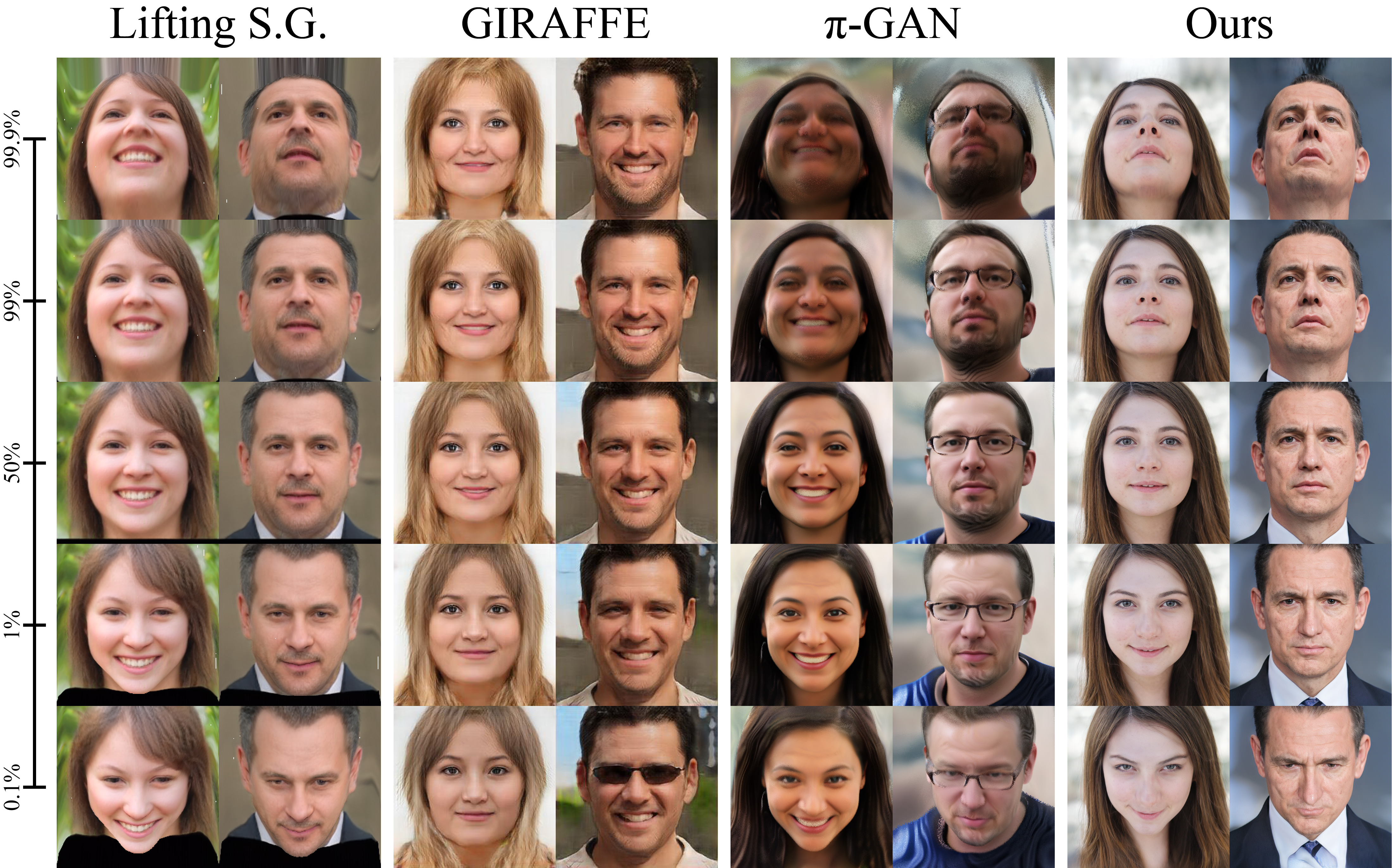}
         \caption{Extrapolation to steep pitch angles.}
         \label{fig:supp_pitch_extrapolation}
     \end{subfigure}
     \hfill
     \begin{subfigure}[b]{0.67\textwidth}
         \centering
         \includegraphics[width=\textwidth]{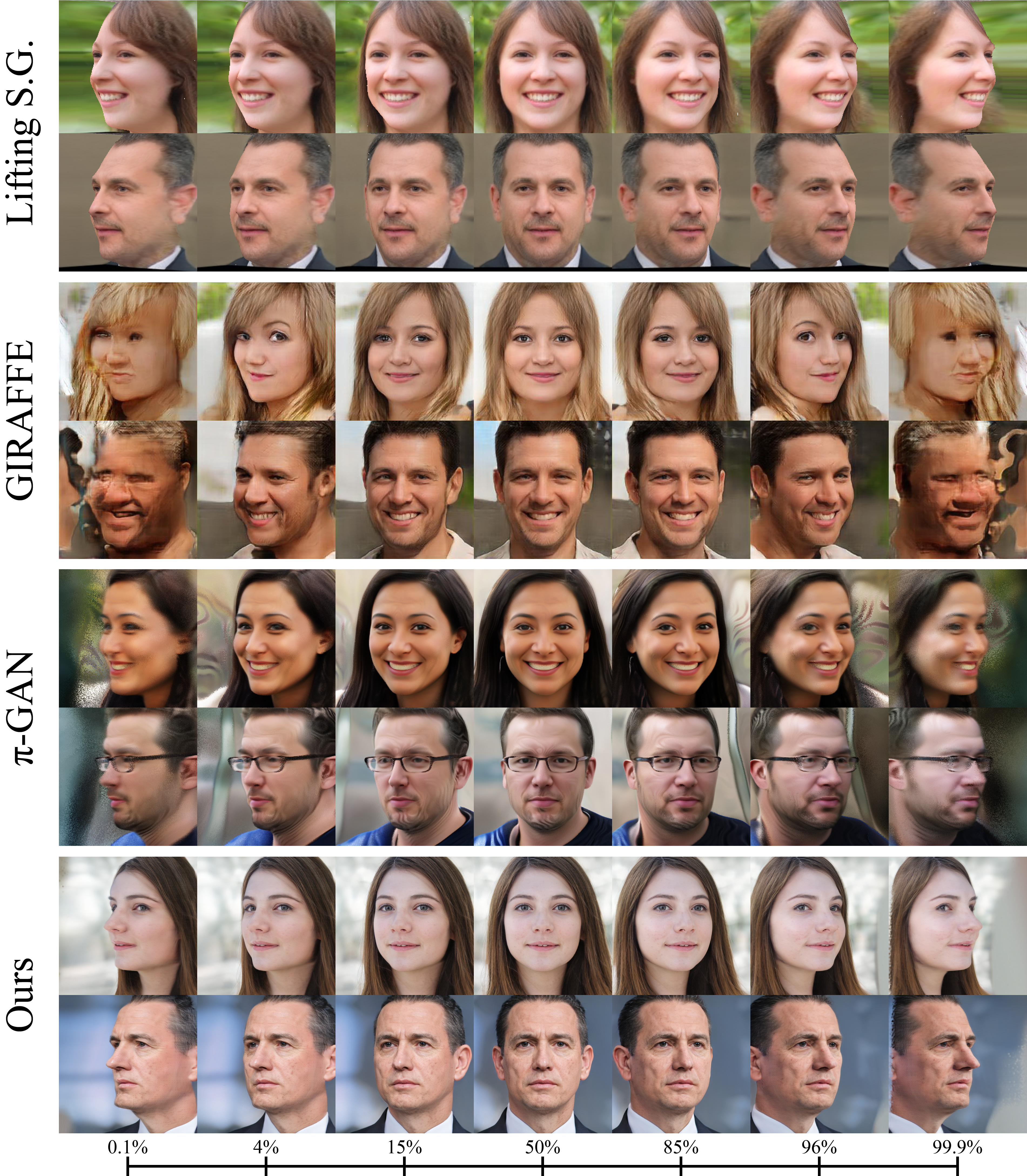}
         \caption{Extrapolation to steep yaw angles.}
         \label{fig:supp_yaw_extrapolation}
     \end{subfigure}
    \caption{We compare methods in their extrapolation to steep camera viewing angles. Labelled is the percentile for camera pitch or yaw. A yaw angle in the 96\textsuperscript{th} percentile means 96\% of training poses are less steep, i.e. 4\% of training poses are beyond the given pose.}
    \label{fig:supp_view_extrapolation}
\end{figure*}

Fig.~\ref{fig:supp_view_extrapolation} provides a visual comparison of our method against baselines for generating views from steep camera poses. We note that the FFHQ\cite{karras2019style} dataset is primarily composed of front-facing images—few images depict faces from extreme yaw angles, and even fewer images depict faces from extreme pitch angles. Nevertheless, reasonable extrapolation to the edges of the pose distribution is a desirable quality and indicates reliance on a robust 3D representation.

Lifting StyleGAN\cite{shi2021lifting}, which represents scenes as a textured mesh, demonstrates consistent rendering quality. However the steep camera angles reveal inaccurate 3D geometry (e.g. foreshortened faces) learned by the method. $\pi$-GAN\cite{chan2020pi}, reasonably extrapolates to steep angles but exhibits visible quality degradation at the edges of the pose distribution.
GIRAFFE\cite{Niemeyer2020GIRAFFE}, being highly reliant on view-inconsistent convolutions, has difficulty reproducing angles that are rarely seen in the dataset. If we force GIRAFFE to extrapolate beyond the camera poses sampled at training (e.g. the leftmost and rightmost images of Fig.~\ref{fig:supp_yaw_extrapolation}), we receive degraded, view-inconsistent images rather than renderings from steeper angles. The problem is amplified for pitch (Fig.~\ref{fig:supp_pitch_extrapolation}) because the dataset's pitch range is even narrower.

Our method, despite also using 2D convolutions, is less reliant on view-inconsistent convolutions for considering the placement of features in the final image. By utilizing an expressive 3D representation as a ``scaffold'', our method provides more reasonable extrapolation to rare views in both pitch and yaw than methods that more strongly depend on image-space convolutions for image synthesis, such as GIRAFFE\cite{Niemeyer2020GIRAFFE}.

\subsection{Additional quantitative results}

Table~\ref{tab:supp_main_results} is an expanded version of Table 2 of the main manuscript that provides additional quantitative metrics, including Kernel Inception Distance~\cite{binkowski2018demystifying} for all datasets and image quality evaluations for ShapeNet Cars. Strong relative performance on Cars, a dataset in which camera poses are distributed uniformly about the sphere,
is evidence that our method is not restricted to face-forward datasets like FFHQ\cite{karras2019style} and AFHQv2\cite{choi2020starganv2, Karras2021}.

\section{Additional visual results}
\label{sec:supp_additional_visual_results}

\begin{figure}
    \centering
    \includegraphics[width=\linewidth]{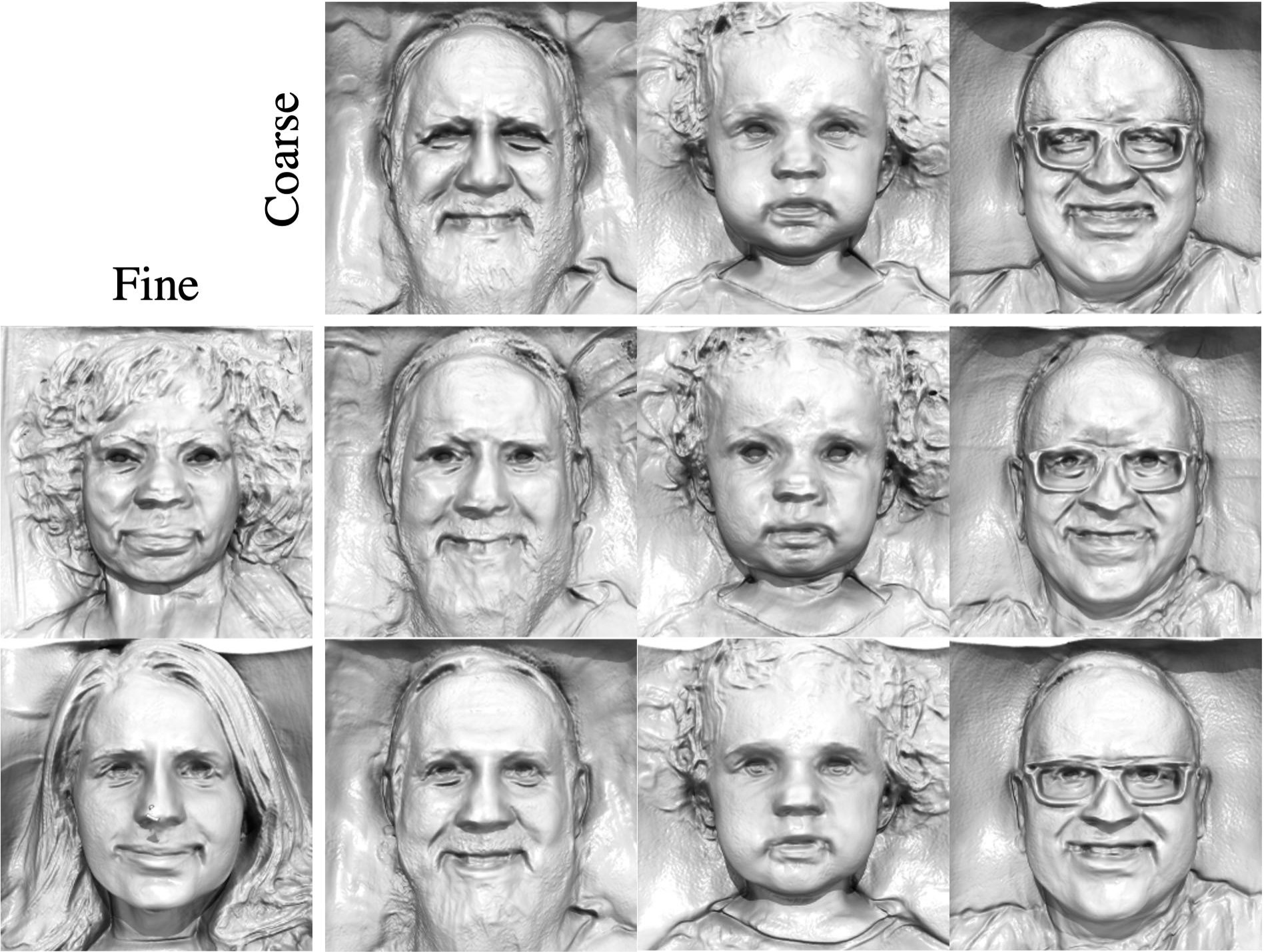}
    \caption{Style-mixing\cite{karras2019style,Karras2020stylegan2,Karras2021} shapes from a model trained on FFHQ $512^2$, without truncation. Aligns with Fig. 8 of the main manuscript, which shows color renderings of the same seeds. The result illustrates that while a mixed example inherits the majority of its structure from its ``coarse'' input (i.e. modulations of layers 0-6), the ``fine'' input (i.e. modulations of layers 7-13) can influence the more delicate details of the shape (e.g. eye regions, hair patterns), in addition to having much control over the overall colors in rendered images.}
    \label{fig:supp_style_mix_shape}
\end{figure}

\paragraph{Style mixing, in shapes.}
Fig.~\ref{fig:supp_style_mix_shape} shows the underlying shapes of the style mixing\cite{karras2019style} examples in Fig. 8 of the main manuscript. While mixed examples inherit most of their shape structure from the modulations of the backbone's low-resolution layers, the modulations of the high-resolution layers can influence fine details in the shape, such as eye regions and hair patterns. The results were obtained from a model trained without style-mixing regularization.


\begin{figure}
     \centering
     \includegraphics[width=\linewidth, keepaspectratio=true]{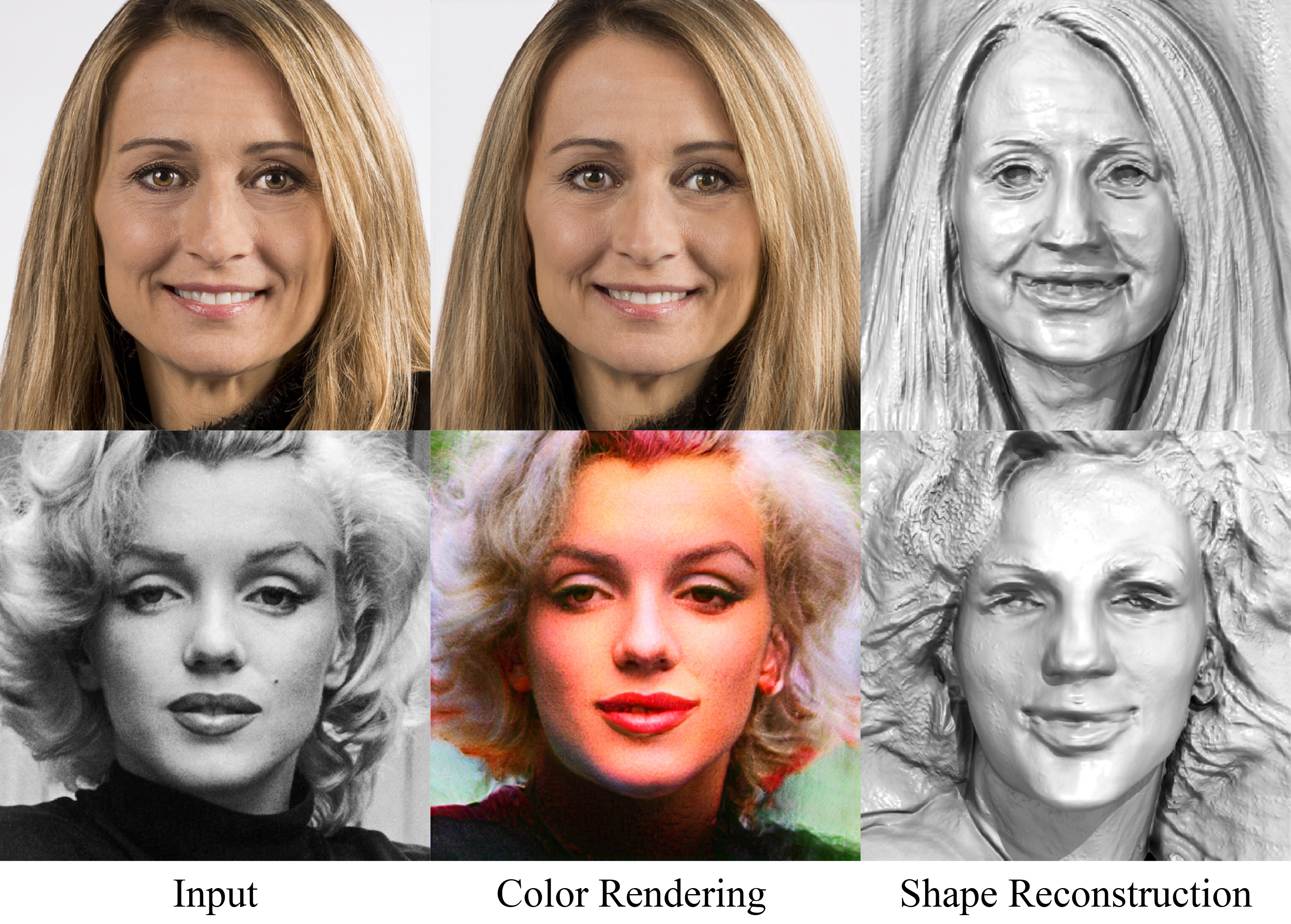}
     \caption{Additional single-view 3D reconstructions of test images demonstrate a use for our generator's  learned prior over facial features.}
     \label{fig:supp_additional_inversions}
\end{figure}

\paragraph{Additional single image 3D reconstructions.}

Fig.~\ref{fig:supp_additional_inversions} provides additional 3D reconstructions of single test images through Pivotal Tuning Inversion (PTI)\cite{roich2021pivotal} of a model trained on FFHQ $512^2$. A pipeline for high-fidelity, single-image reconstruction of faces that does not require explicit 3D ground-truth training data opens the door for many promising applications, such as photo-to-avatar creation.

\begin{figure*}
    \centering
    \includegraphics[width=\textwidth]{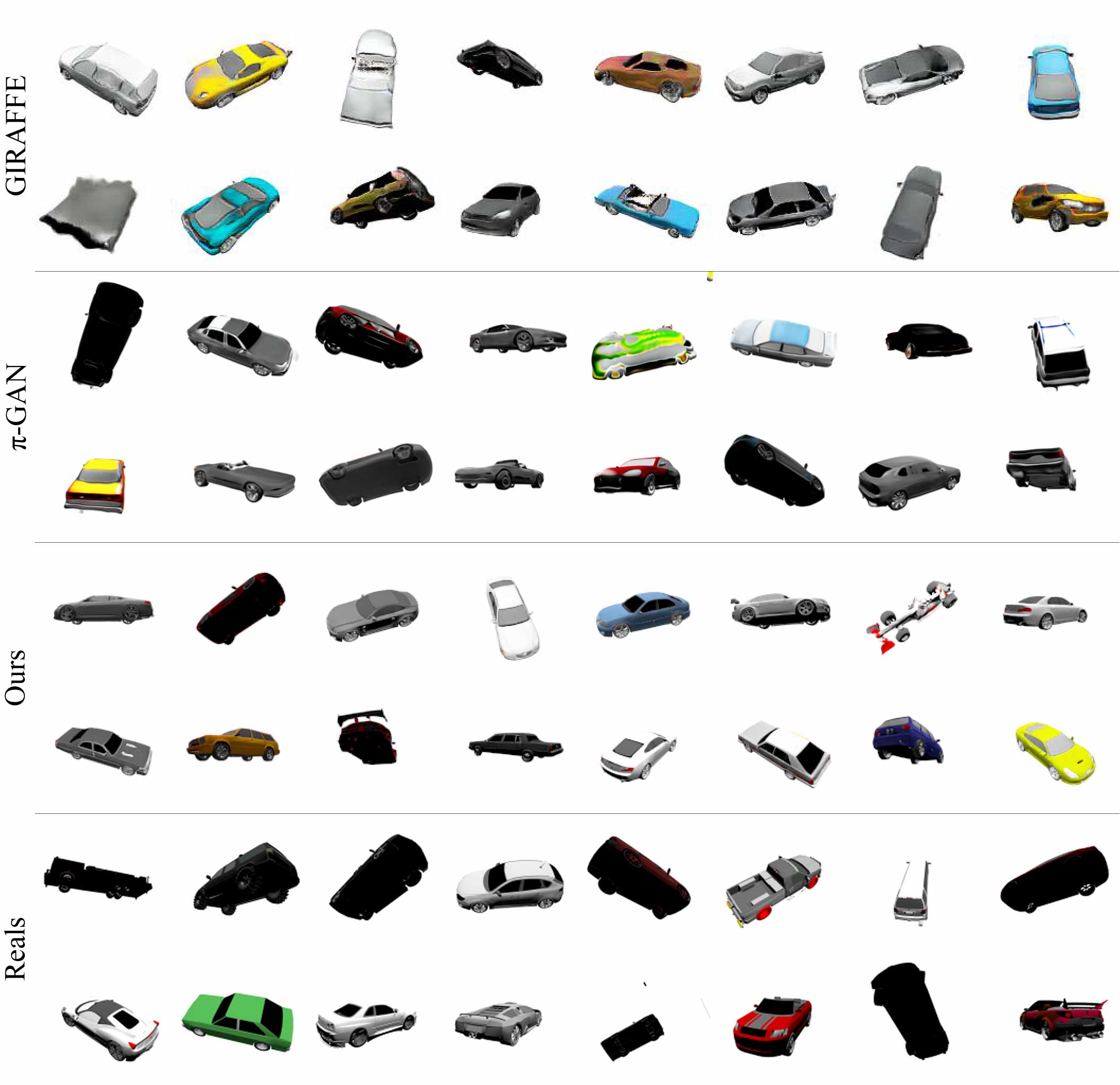}
    \caption{Qualitative comparison of uncurated examples of cars. All methods are sampled with truncation\cite{marchesi2017megapixel, brock2018large, karras2019style}, using $\psi=0.7$.}
    \label{fig:supp_uncurated_cars_comparison}
\end{figure*}

\paragraph{Shapenet Cars.}

Fig.~\ref{fig:supp_uncurated_cars_comparison} contains uncurated renderings from random camera poses for models trained with ShapeNet Cars\cite{chang2015shapenet, sitzmann2019srns}. This experiment serves as a demonstration that our method is capable of operating successfully on datasets that include camera poses that span the entire 360$^{\circ}$ camera azimuth and 180$^{\circ}$ camera elevation distributions, unlike 2.5D GANs\cite{shi2021lifting}, which are intended for face-forward datasets.

\begin{figure*}
    \centering
    \includegraphics[width=\linewidth, keepaspectratio=true]{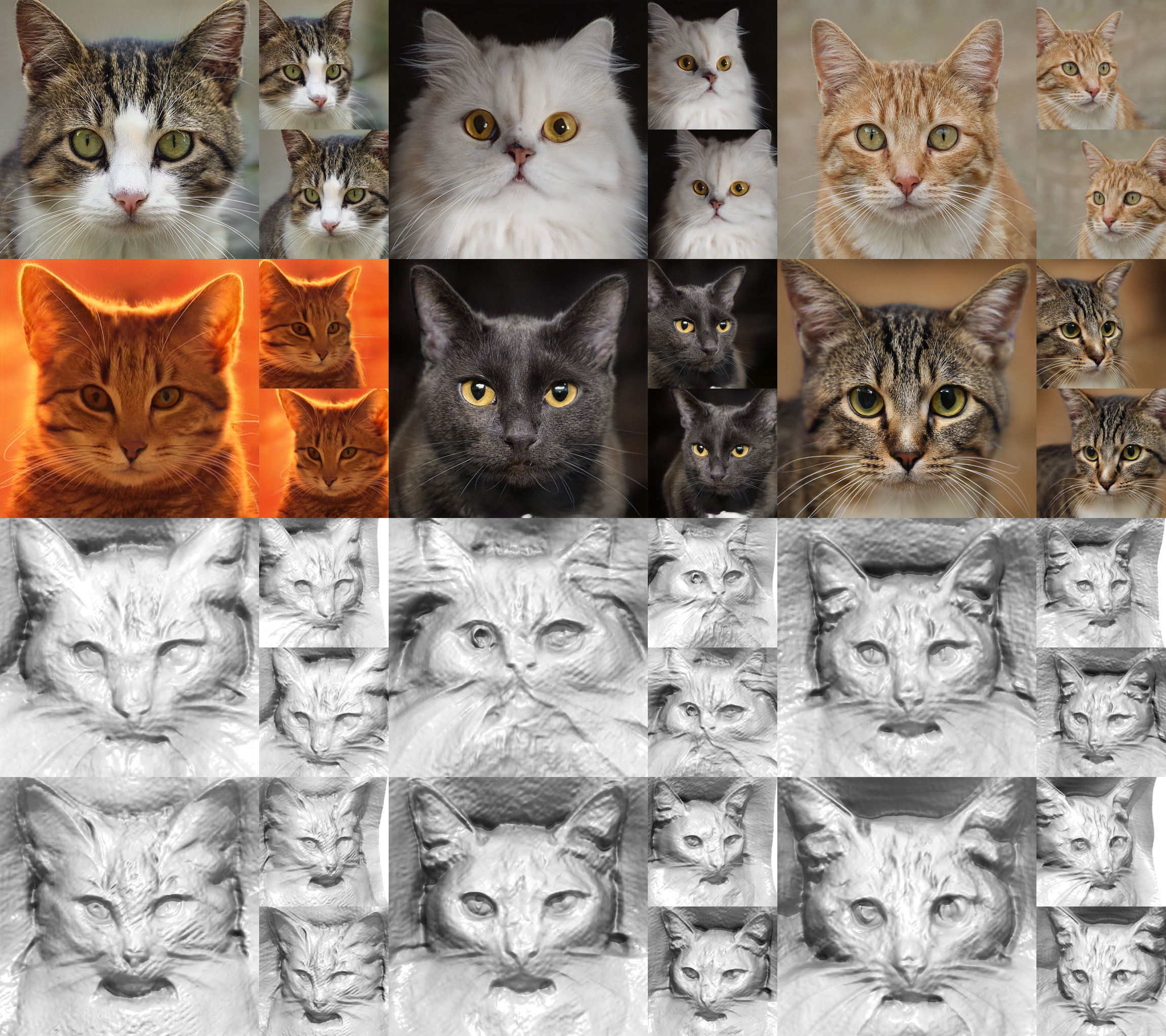}
    \caption{Curated examples from a model trained on AFHQv2\cite{choi2020starganv2,Karras2021} $512^2$.}
    \label{fig:supp_curated_afhq}
\end{figure*}

\paragraph{Additional selected examples synthesized with AFHQv2 Cats.}

Fig.~\ref{fig:supp_curated_afhq} shows renderings and shapes for selected examples, synthesized by our method trained on AFHQv2 Cats\cite{choi2020starganv2, Karras2021} $512^2$.

\begin{figure*}
    \centering
    \includegraphics[width=\textwidth]{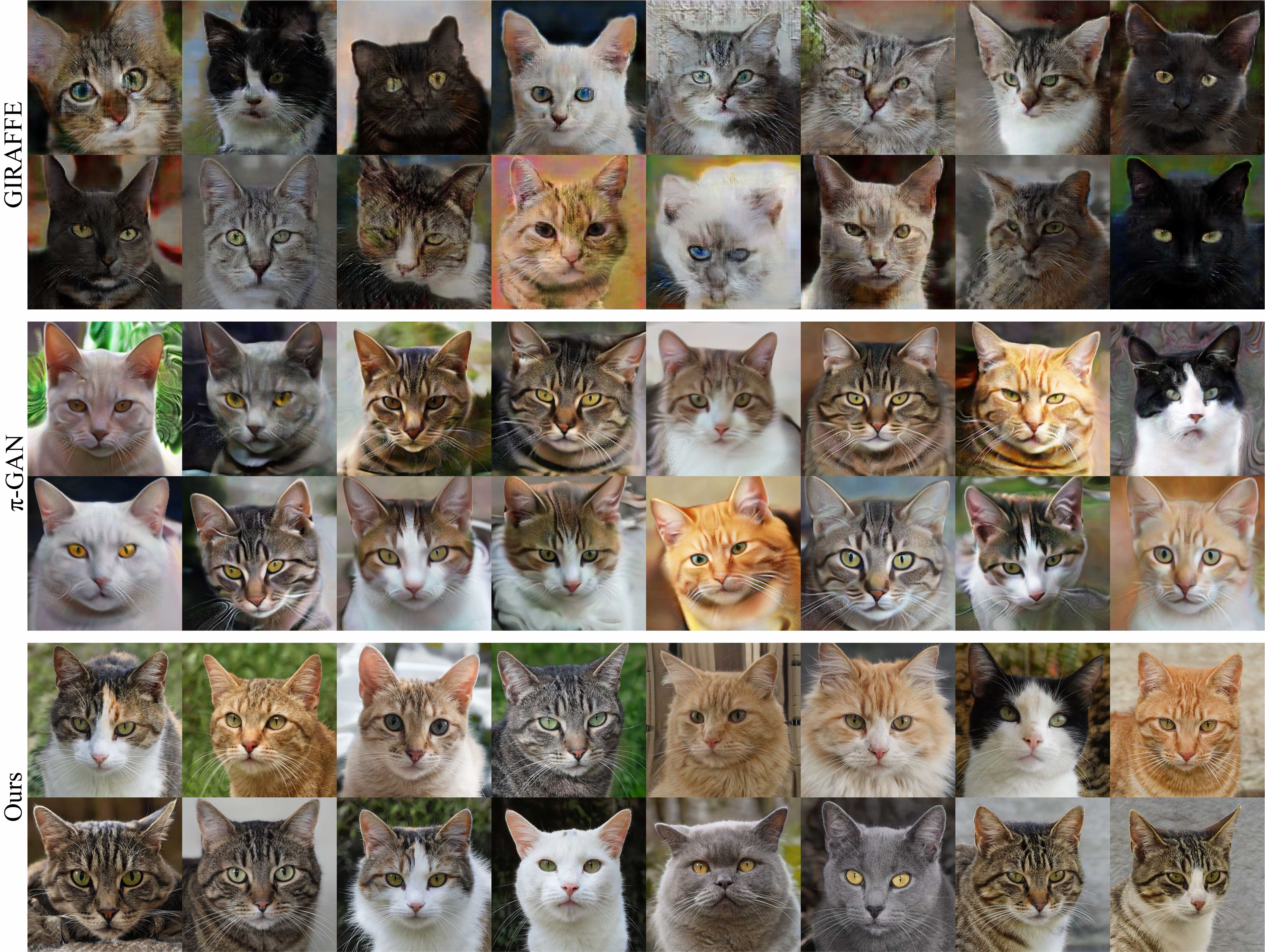}
    \caption{Uncurated examples of cats, for GIRAFFE\cite{Niemeyer2020GIRAFFE} $256^2$, $\pi$-GAN $128^2$, and our method $512^2$. All methods are sampled with truncation\cite{marchesi2017megapixel,brock2018large,karras2019style}, using $\psi=0.7$.}
    \label{fig:uncurated_cats_comparison}
\end{figure*}

\paragraph{Uncurated examples synthesized with AFHQv2 Cats.}

Fig.~\ref{fig:uncurated_cats_comparison} provides uncurated examples of cats produced by GIRAFFE\cite{Niemeyer2020GIRAFFE}, $\pi$-GAN\cite{chan2020pi}, and our method, trained at image rsolutions of $256^2$, $128^2$, and $512^2$, respectively.

\begin{figure*}
    \centering
    \includegraphics[width=\textwidth]{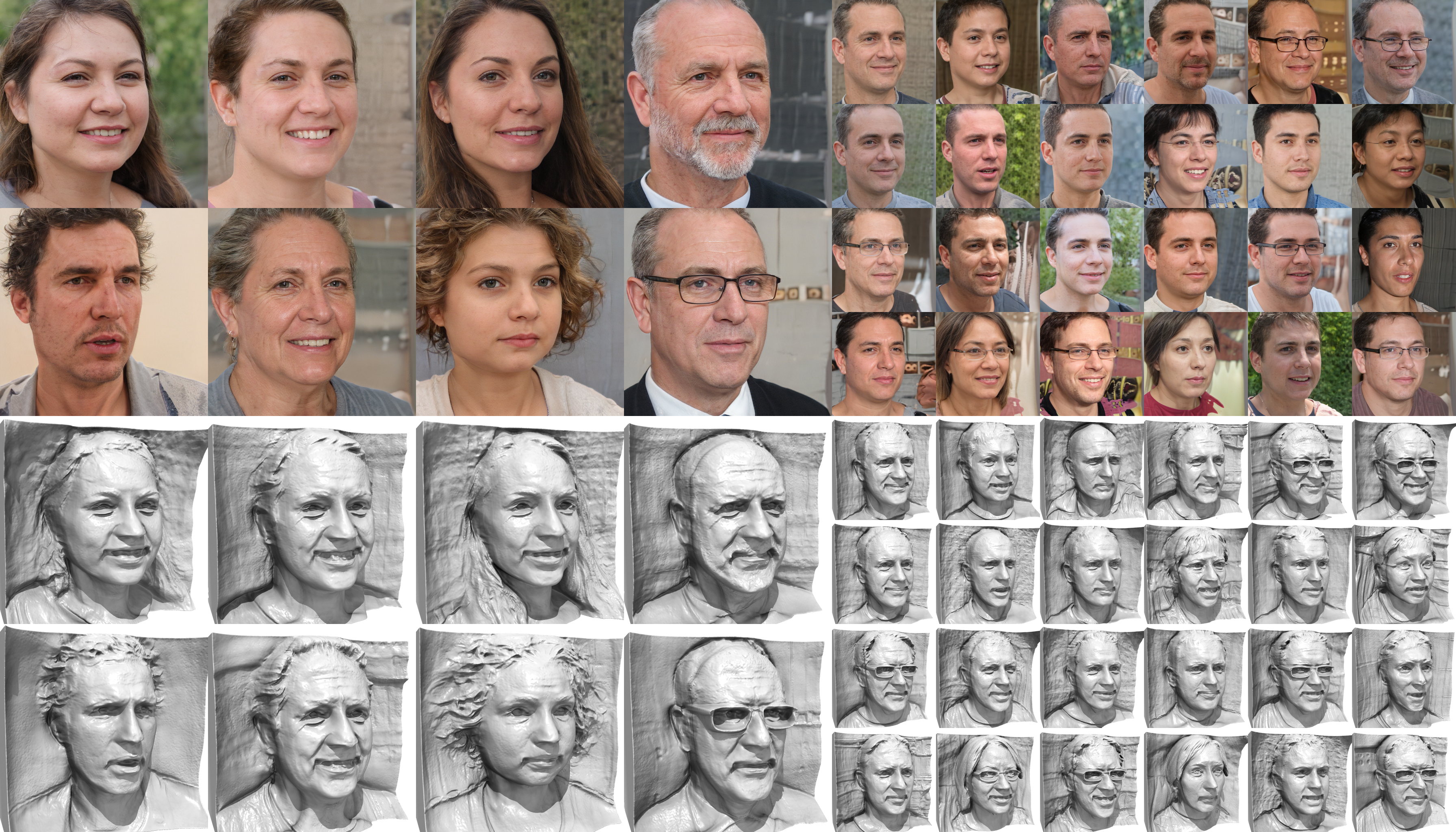}
    \caption{Images and geometry for seeds 0-31, synthesized using a model trained on FFHQ\cite{karras2019style} $512^2$. Sampled with truncation\cite{karras2019style}, using $\psi=0.5$.}
    \label{fig:supp_uncurated_ffhq}
\end{figure*}

\paragraph{Uncurated examples synthesized with FFHQ.}

Fig.~\ref{fig:supp_uncurated_ffhq} provides uncurated examples of faces produced by our method, trained with FFHQ\cite{karras2019style} $512^2$. We apply truncation\cite{marchesi2017megapixel, brock2018large, karras2019style}, with $\psi=0.5$.


\begin{figure*}
    \centering
    \includegraphics[width=\textwidth]{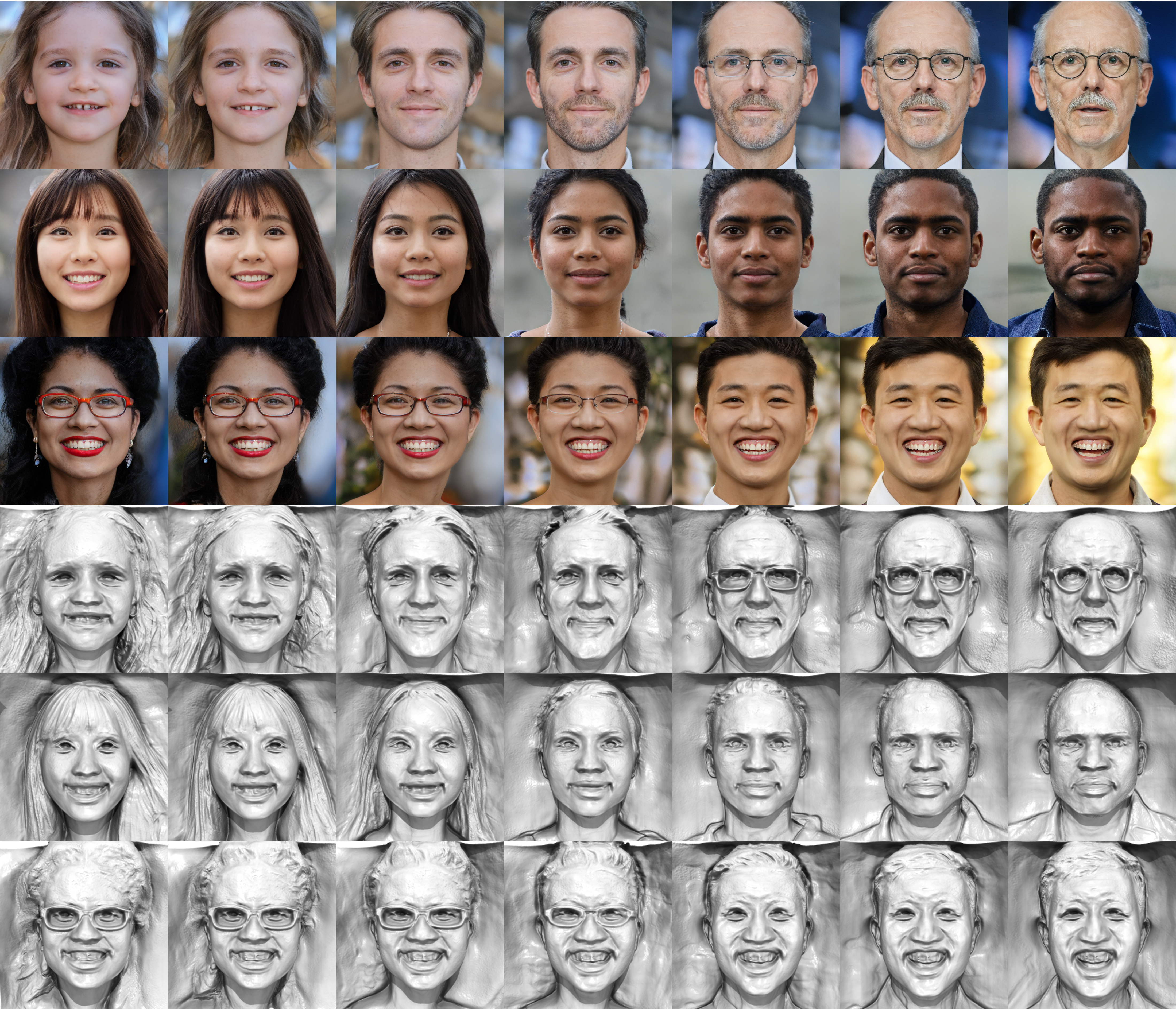}
    \caption{Linear interpolations between latent codes, showing renderings and shapes.
    }
    \label{fig:supp_interpolation}
\end{figure*}

\paragraph{Latent code interpolation.}

Fig.~\ref{fig:supp_interpolation} provides linear interpolations between latent codes for selected examples produced by our method trained on FFHQ $512^2$. Our result illustrates that our 3D GAN inherits the well-behaved latent space of the StyleGAN2\cite{Karras2020stylegan2} backbone, which enables smooth interpolations in both color renderings and underlying shapes.

\begin{figure*}
    \centering
    \includegraphics[width=\textwidth, keepaspectratio=true]{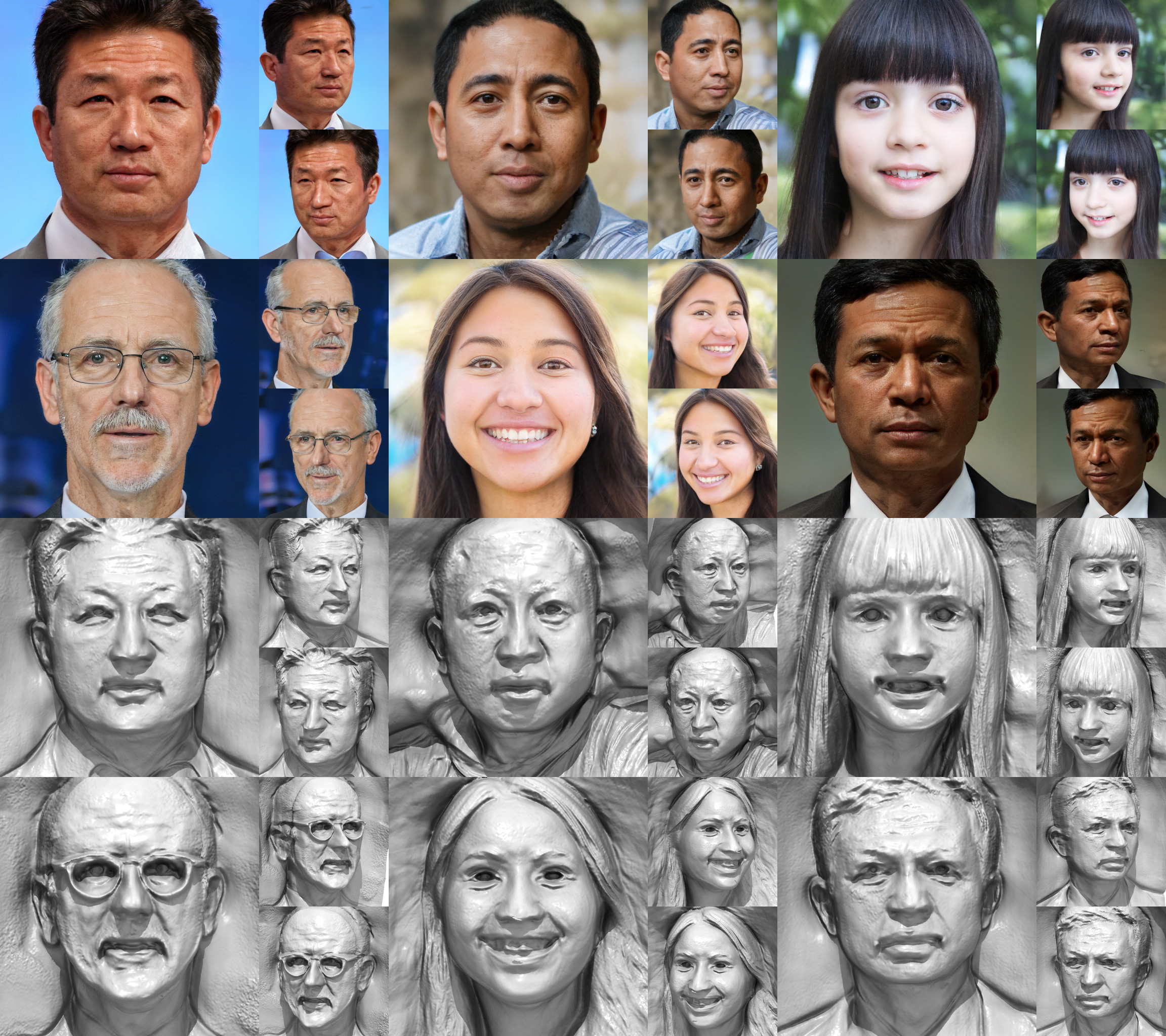}
    \caption{Additional selected examples, from a model trained on FFHQ\cite{karras2019style} $512^2$.}
    \label{fig:supp_curated_ffhq}
\end{figure*}

\paragraph{Additional selected examples synthesized with FFHQ}

Fig.~\ref{fig:supp_curated_ffhq} depicts renderings and shapes for selected examples, synthesized by our method trained on FFHQ $512^2$.

\section{Implementation details}
\label{sec:supp_implementation_details}

\begin{figure*}
    \centering
    \includegraphics[width=\textwidth]{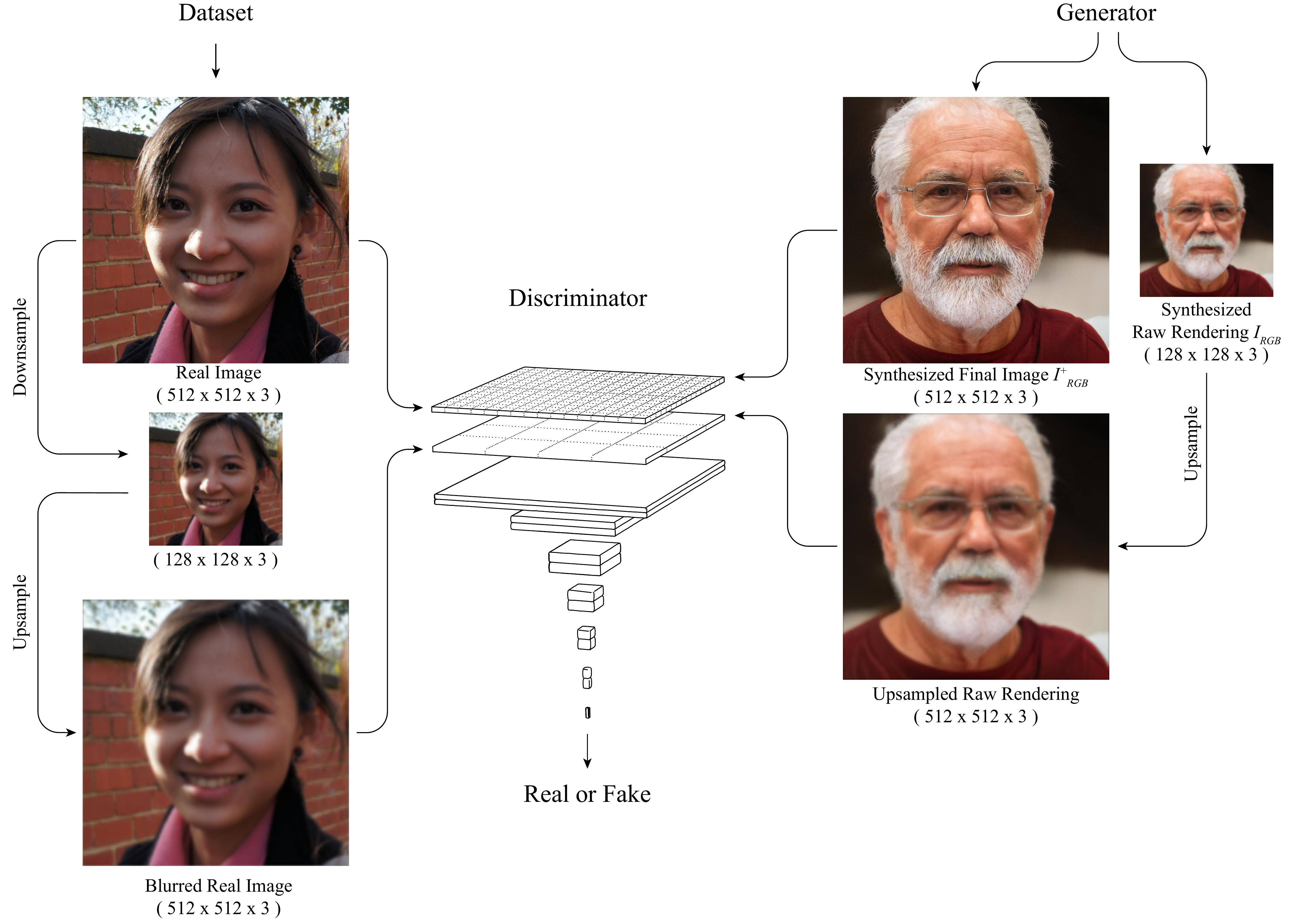}
    \caption{In dual-discrimination, we discriminate on a six-channel concatenation of the final image and the raw neural rendering, in order to maintain consistency between high-resolution final images and view-consistent (but low resolution) neural renderings. This diagram illustrates how we obtain a six-channel discriminator input tensor for both real and fake images. Our generator produces both a $512^2$ final rendering ($I^+_{RGB}$) as well as the ($128^2$) raw neural rendering ($I_{RGB}$). The raw rendering, $I_{RGB}$ is the first three channels of the 32-channel rendered features, $I_F$. We create a six-channel discriminator input by upsampling the raw image to $512^2$ and concatenating it with the final image to form a ($512 \times 512 \times 6$) discriminator input tensor. For real images, we extract a $512^2$ real image from the dataset and downsample it to the same size as $I_{RGB}$ to obtain an analogue for $I_{RGB}$. We then upsample this image back to $512^2$ and concatenate it with the original image to form a ($512 \times 512 \times 6$) discriminator input tensor. The downsample-then-upsample operation has the effect of blurring the original image.
    }
    \label{fig:supp_dd-in-detail}
\end{figure*}

We implemented our 3D GAN framework on top of the official PyTorch implementation of StyleGAN2, an updated version of which is available at \url{https://github.com/NVlabs/stylegan3}. Most of our training parameters are identical to those of 
StyleGAN2
\cite{Karras2020stylegan2}, including the use of equalized learning rates for the trainable parameters\cite{karras2018progressive}, a minibatch standard deviation layer at the end of the discriminator\cite{karras2018progressive}, exponential moving average of the generator weights, and a non-saturating logistic loss\cite{goodfellow2014generative} with R1 regularization\cite{gan_convergence}.


\paragraph{Two-stage training.}
In order to save computational resources, we perform the majority of the training at a neural rendering resolution of $64^2$, before gradually stepping the resolution up to $128^2$. Note that the final image resolution remains fixed throughout training (e.g. $256^2$ or $512^2$). We implement this simply by bilinearly resizing the raw neural rendering $I_{RGB}$ to $128^2$ before it is operated on by the super-resolution module. Thus, the super-resolution module always receives a $128^2$-sized feature map as an input, regardless of the actual neural rendering resolution. In contrast to previous progressive growing strategies\cite{karras2018progressive, chan2020pi} that double the resolution in a single step, we gradually increase the neural rendering resolution, pixel-by-pixel, over 1 million images, i.e., ($64^2$, $65^2$, $66^2$, ..., $126^2$, $127^2$, $128^2$). We continue training with the resolution fixed at $128^2$ for an additional 1.5 million images, for a total of 2.5M iterations of fine-tuning.
This two-stage training procedure provides a roughly $2\times$ speed-up versus training from scratch at full resolution and produces similar results to training at full neural rendering resolution from scratch.

\paragraph{Backbone.}
Our backbone (i.e., StyleGAN2 generator) follows the implementation of \cite{Karras2020stylegan2}, with a mapping network of 8 hidden layers. For all of our experiments (regardless of final image resolution), the backbone operates at a resolution of $256^2$. We modify the output convolutions such that they produce a 96-channel output feature image, which we reshape into three planes, each of shape $256 \times 256 \times 32$. Unlike approaches that require pre-trained 2D image GANs\cite{shi2021lifting}, we do not utilize pre-trained StyleGAN2 checkpoints for the backbone; the entire pipeline is trained end-to-end. For large datasets, such as FFHQ\cite{karras2019style} and ShapeNet Cars\cite{chang2015shapenet, sitzmann2019srns}, we train from scratch with random initialization; for small datasets, such as AFHQv2\cite{choi2020starganv2, Karras2021}, we follow prevailing methodology\cite{Karras2020ada} by fine-tuning from a checkpoint trained on a larger dataset.

\paragraph{Decoder and volume rendering.}
Our decoder is implemented as an MLP with a single hidden layer of 64 hidden units and uses the softplus activation function. The decoder takes as input a 32-channel aggregated feature vector; it produces a 33-channel vector that we split into a scalar density prediction and a 32-channel feature. We use neural volume rendering~\cite{mildenhall2020nerf} of features\cite{Niemeyer2020GIRAFFE}, with two-pass importance sampling. For FFHQ~\cite{karras2019style} and AFHQv2~\cite{choi2020starganv2, Karras2021}, we use 48 uniformly-spaced and 48 importance samples per ray; for ShapeNet Cars, we use 64 uniformly-spaced and 64 importance samples per ray. When rendering videos that feature thin surfaces, we found it beneficial to increase the samples per ray during inference to reduce flicker.

\paragraph{Super-resolution.}
We implement our super-resolution model with two `blocks' of StyleGAN2's modulated convolutions\cite{Karras2020stylegan2}, with noise inputs disabled. The blocks contain convolutions of channel-depth 128 and 64, respectively.

\paragraph{Discriminator.}
Our discriminator is a StyleGAN2\cite{Karras2020stylegan2} with two modifications. First, to enable dual discrimination, we adjust the input layer to accept six-channel input images, rather than 3-channel input images. Fig.~\ref{fig:supp_dd-in-detail} provides a diagram that illustrates the creation of these six-channel inputs, for both real and generated images. Second, we condition the discriminator on the camera parameters of the incoming image to help prevent degenerate shape solutions; we follow the class-conditional discriminator modifications of \cite{Karras2020ada} to inject this information.

\paragraph{Mixed Precision.}
To speed up training, we use a similar mixed-precision methodology as \cite{Karras2020ada}. We use FP16 in the four highest resolution blocks of the discriminator and in both blocks of our super-resolution module. We do not use FP16 in our generator backbone.

\paragraph{R1 Regularization.}

We use R1 regularization\cite{gan_convergence} with $\gamma=1$ for all datasets and resolutions, except for ShapeNet Cars, where we use $\gamma=0.1$. Regularization strengths were informally
chosen based on values that have shown success with previous methods\cite{Karras2020stylegan2, Karras2020ada}.

\paragraph{Density Regularization.}

Further experiments, conducted after our initial submission, suggested that additional regularization over the estimated density field reduced the prevalence of undesirable seams and other shape artifacts. Similar to the total variation regularization used in previous work \cite{Lombardi:2019}, our density regularization encourages smoothness of the density field. For each generated scene in the batch, we randomly sample points $\mathbf{x}$ in the volume and also sample additional `perturbed' points that are offset with a small amount Gaussian noise $\delta\mathbf{x}$. Our density regularization loss is an L1 loss that minimizes the difference between the estimated densities $\sigma(\mathbf{x})$ and $\sigma(\mathbf{x}+\delta\mathbf{x})$.
We apply our density regularization over 1000 pairs of randomly sampled points every four training iterations.
\paragraph{Training.}
We train all models with a batch size of 32. We use a discriminator learning rate of 0.002 and a generator learning rate of 0.0025. Following \cite{Karras2021}, we blur images as they enter the discriminator, gradually reducing the blur amount over the first 200K images. Unlike \cite{Karras2020stylegan2}, we train without style-mixing regularization.


Using the two-stage training discussed previously, we train at a resolution of $64^2$ for 25M images and at $128^2$ for an additional 2.5M images.
Using a neural rendering resolution of $64^2$, our 3D GAN framework takes $\sim$24 seconds to train on 1000 images (24 s/kimg) on 8 Tesla V100 GPUs; this increases to $46$ s/kimg at a neural rendering resolution of $128^2$. For reference, StyleGAN3-R\cite{Karras2021} achieves training rates of $20$ s/kimg on similar hardware.

Our total training time on 8 Tesla V100 GPUs is on the order of 8.5 days (7 days of $64^2$ training, plus 1.5 days of $128^2$ fine-tuning), compared to 6 days on similar hardware for StyleGAN3-R. 

\paragraph{Inference-time depth samples.}
We use neural volume rendering~\cite{mildenhall2020nerf} with two-pass importance sampling to render feature images from our tri-plane representation. We found that increasing the number of samples per ray at inference time can reduce unwanted flickering when rendering videos that feature thin objects such as eye glasses. For clips shown in the supplemental video, we double both the number of coarse samples (from 48 to 96) and the number of fine samples (from 48 to 96), bringing the total number of depth samples per ray to 192. Increasing the number of samples per ray incurs a penalty to the rendering speed. When using 96 total depth samples per ray, frame rates are reduced to approximately 24 frames per second with tri-plane caching -- down from 36 frames per second when using the default 48 samples. Images shown in the main manuscript were synthesized without increasing the number of depth samples along each ray.

\paragraph{AFHQv2.}
Following \cite{Karras2020ada}, we fine-tune from FFHQ-trained models to achieve optimum performance on Cats. Beginning from a checkpoint trained on FFHQ, we train for 6.2M images at a neural rendering resolution of $64^2$; and for an additional 2.6M images, while fine-tuning the neural rendering resolution to $128^2$. Because $\pi$-GAN and GIRAFFE were not designed with the benefits of adaptive discriminator augmentation (ADA) \cite{Karras2020ada}, we also do not use ADA for our method at $256^2$, in an effort to keep comparisons across methods fair. We use adaptive discriminator augmentation with its default settings, for our method only at $512^2$ .





\section{Experiment details}
\label{sec:supp_experiment_details}

\subsection{Baselines}
$\pi$-GAN\cite{chan2020pi} is a 3D-aware GAN that relies upon a FiLM-conditioned MLP with periodic activation functions for camera-controllable synthesis. We utilized the official code (\url{https://github.com/marcoamonteiro/pi-GAN}) and trained until convergence with the parameters recommended for analogous datasets.

GIRAFFE\cite{Niemeyer2020GIRAFFE} is a 3D-aware GAN that incorporates a compositional 3D scene representation to enable controllable synthesis. We utilized the official code (\url{https://github.com/autonomousvision/giraffe}) and trained until convergence with the parameters recommended for analogous datasets.

Lifting StyleGAN\cite{shi2021lifting} is a method for disentangling and lifting a pre-trained StyleGAN2 image generator to 3D-aware face generation. The original Lifting StyleGAN manuscript reports results on a slightly tighter crop of FFHQ than we used. Because we had difficulty matching the quality of Lifting StyleGAN's pre-trained model when we trained it from scratch on our less-cropped dataset, we instead used their official pre-trained model for their tighter crops and the FID score reported in their manuscript. We utilized the offical code, found here: (\url{https://github.com/seasonSH/LiftedGAN}).

StyleGAN2 is a style-based GAN that achieves state-of-the-art image quality for 2D image synthesis and features a well-behaved latent space that enables image manipulation. We obtained a pre-trained checkpoint for StyleGAN2 on FFHQ $512^2$ from the collection of official models (\url{https://catalog.ngc.nvidia.com/orgs/nvidia/teams/research/models/stylegan2}). Following the recommended tuning of \cite{Karras2020ada}, we trained both StyleGAN2 config F and the $512 \times 512$ config from \cite{Karras2020ada}, sweeping R1~\cite{gan_convergence} regularization strength, $\gamma = \{0.2, 0.5, 1, 2, 5, 10, 20 \}$. The best result for AFHQv2 was obtained with StyleGAN2 config F, after training for 10M images at $\gamma=1$. 




\subsection{Dataset Details}

\paragraph{FFHQ}
We prepare our dataset by starting with the "in-the-wild" version of the FFHQ dataset \cite{karras2019style}, which is composed of uncropped, original PNG images of people sourced from Flickr. We use an off-the-shelf face detection and pose-extraction pipeline\cite{deng2019accurate} to both identify the face region and label the image with a pose. We crop the images to roughly the same size as the original FFHQ dataset.

We assume fixed camera intrinsics across the entire dataset, with a focal length of $4.26 \times image\_width$, equivalent to a standard portrait lens. We prune a small number of images that resisted face detection; our final dataset contains 69957 images. We augment the dataset with horizontal flips.

\paragraph{AFHQv2}
We used the AFHQv2 dataset \cite{Karras2021}, which is a higher-quality version of the original AFHQ dataset \cite{choi2020starganv2}. AFHQv2 provides closeups for animal faces including cats, dogs, and wildlife. We use the `cats' split, which contains approximately 5000 images, for our experiments. As with FFHQ, we assume fixed camera intrinsics across the dataset; for simplicity, we use identical intrinsics to FFHQ. Camera poses were extracted via landmark detection\cite{cat_hipster} and an open-source Perspective-n-Point algorithm\cite{opencv_library}. We augment the dataset with horizontal flips.


\paragraph{ShapeNet Cars}
For additional validation, we compare methods on ShapeNet Cars\cite{chang2015shapenet, sitzmann2019srns} to evaluate performance on a dataset that contains views from all angles. We adopted the dataset and setup from \cite{sitzmann2019srns}, which is composed of $128^2$ resolution renderings of synthetic cars, each labelled with camera parameters. The dataset contains 2457 unique cars; each car is rendered from 50 views randomly sampled from the entire sphere. We use the known camera parameters for each image and do not augment the dataset with image space augmentations.


\subsection{Single scene overfitting.}

To illustrate the effectiveness of our architecture, we evaluate the relative performance of the tri-plane 3D representation against a comparable voxel-based hybrid representation and Mip-NeRF\cite{barron2021mipnerf} on the \textit{Family} scene of Tanks \& Temples\cite{Knapitsch2017} dataset as desribed in Section 3 of the main manuscript. We use the pre-processed images, as well as the training/test split, of \cite{liu2020neural}. We use 512 uniformly-spaced depth samples and 256 importance samples per ray and a ray batch size of 6400. The tri-planes are treated as learnable parameters of shape $3 \times 48 \times 512 \times 512$. The dense voxel parameters were chosen to optimize quality for comparable parameter count as the tri-planes; the voxel features are of shape $18 \times 128 \times 128 \times 128$. Both voxel and tri-plane hybrid representations are coupled with two-layer, 128 hidden unit decoders with Fourier feature embeddings\cite{tancik2020fourier}. We train voxel and cube representations for 200K iterations; we train Mip-NeRF for the recommended 1M iterations.



\subsection{Pivotal tuning inversion.}

We use off-the-shelf face detection~\cite{deng2019accurate} to extract appropriately-sized crops and camera extrinsics from test images and we resize each cropped image to $512^2$. We follow Pivotal Tuning Inversion (PTI)\cite{roich2021pivotal}, optimizing the latent code for 500 iterations, followed by fine-tuning the generator weights for an additional 500 iterations.

For inversion of grayscale images, we convert the generator's 3-channel, $RGB$ renderings to perceived luminance,  $Y$, before computing image distance loss during optimization. This allows the generator's prior to colorize the renderings. To compute single-channel luminance from 3-channel $RGB$ images, we use $Y = 0.299R + 0.587G + 0.114B$. For grayscale optimization, we use 400 latent code inversion steps and 250 generator fine-tuning steps.


\subsection{Evaluation Metrics}

\paragraph{FID and KID.} We compute Fr\'echet Inception Distance ({FID})~\cite{DBLP:journals/corr/HeuselRUNKH17} and Kernel Inception Distance (KID)\cite{binkowski2018demystifying} image quality metrics between 50k generated images and all training images using the implementation provided in the StyleGAN3\cite{Karras2021} codebase.



\paragraph{Geometry.}
We follow a similar procedure to \cite{shi2021lifting} in the evaluation of geometry. We generate 1024 images and depth maps from random poses that match the dataset pose distribution. With the application of a pre-trained 3D face reconstruction model\cite{deng2019accurate}, we generate a ``pseudo'' ground-truth depth map for each generated image. Next we limit both the generated depth maps and ``pseudo'' ground-truth depth maps to the facial regions as defined by the reconstruction model. Finally, we normalize all depth maps to zero mean, unit variance and calculate the L2 distance between them.

\paragraph{Multi-view consistency.}

We evaluate multi-view consistency and face identity preservation for models trained on FFHQ~\cite{karras2019style} by measuring ArcFace~\cite{deng2018arcface} cosine similarity. For each method, we generate 1024 random faces and render two views of each face from poses randomly selected from the training dataset pose distribution. For each image pair, we measure facial identity similarity~\cite{deng2018arcface} and compute the mean score.

\paragraph{Pose accuracy.}

We evaluate pose accuracy with the help of a pre-trained face reconstruction model\cite{deng2019accurate}. With~\cite{deng2019accurate}, we detect pitch, yaw, and roll from 1024 generated images then compute L2 loss against the ground truth poses to determine each model's pose drift.

\paragraph{Runtime.}
We evaluate runtime for each model by calculating the average framerate over a 400 frame sequence. We process frames consecutively, i.e., with batch size 1. In order to give each method a best-case-scenario, we ignore operations such as copying rendered frames from GPU to CPU and saving files to disk.

\paragraph{FACS estimation}
In Section 5.2 of the main paper, we quantitatively measure the effect of dual discrimination and generator pose conditioning at preserving facial expressions across multi-view face videos. To evaluate facial expressions, we employ a proprietary facial tracker that measures detailed movement of sub-regions of the face in terms of Facial Action Coding System (FACS)~\cite{Ekman1978} coefficients. Specifically, our facial tracker measures all 53 FACS blendshape coefficients defined in Li et al.~\cite{li2020learning} and we compared the variability in the `mouthSmile\_L' and `mouthSmile\_R' blendshape coefficients across the different videos.

\subsection{Visualization of Geometry}

To visualize shapes, we sample the volume to obtain a $512^3$ cube of density values and extract the surface of the scene as a mesh using Marching Cubes~\cite{lorensen1987marching}. We found that a levelset between 0 and 10 generally yielded visually appealing results. Renderings of shapes shown in this manuscript were generated using ChimeraX~\cite{goddard2018ucsf}.

\section{Discussion}
\label{sec:supp_discussion}

\subsection{Shape artifacts}


Despite significant improvements in the quality of the 3D geometry compared to previous methods, our synthesized shapes are not free from artifacts, which are visible in geometry renderings throughout the main paper and supplement (e.g. Fig.~\ref{fig:supp_uncurated_ffhq}, Fig.~\ref{fig:supp_curated_ffhq}). Sunken eye sockets allow the illusion of eyes that follow the viewing camera, even when the geometry and neural renderings are view-consistent; such ``hollow face illusions" have demonstrated similar effects in the physical world. Similarly, deep creases near the corners of mouths enable the creation of ``view-inconsistent'' effects that in fact are faithful to the underlying shapes. Future work that incorporates stronger dataset priors, e.g. that eyeballs are convex, may help resolve these artifacts.

While our method produces more-detailed eyeglasses than previous methods, it tends to produce ``goggles''—the sides of the eyeglasses are opaque where there should be empty space. Future neural rendering methods that can accurately model lens refraction may enable more faithful reconstruction of eyeglasses and other objects that contain transparent elements.

In some shapes and renderings generated by our method, a seam is visible between the face and the rest of the head. While we find the optional density regularization in Sec. \ref{sec:supp_implementation_details} helps reduce such artifacts, we hypothesize that recent hybrid-SDF rendering solutions\cite{oechsle2021unisurf, wang2021neus, yariv2021volume}, which have shown promising results in robust geometry recovery from images, may yield improved shapes with fewer artifacts.

In the interests of simplicity, we model the scene with a single 3D representation, without any explicit background handling. Consequently, the generator learns to represent backgrounds of images with textured surfaces fused to foreground objects. Future work that models backgrounds with a separate 3D representation~\cite{kaizhang2020, Niemeyer2020GIRAFFE, niemeyer2021campari} may enable isolation of foreground objects. 

\clearpage

{\small
\bibliographystyle{ieee_fullname}
\bibliography{egbib}
}